\documentclass[lettersize,journal]{IEEEtran}

\usepackage{multirow}
\usepackage{booktabs}
\usepackage{graphicx}
\usepackage[normalem]{ulem}

\useunder{\uline}{\ul}{}

\usepackage{amsmath,amsfonts}
\usepackage{algorithm}
\usepackage{array}
\usepackage{subfig}
\usepackage{textcomp}
\usepackage{stfloats}
\usepackage{url}
\usepackage{verbatim}
\usepackage{graphicx}
\usepackage{cite}
\usepackage{amssymb}
\usepackage{xcolor}
\usepackage{pifont}
\usepackage{algpseudocode}
\usepackage{color}
\usepackage{ulem}
\usepackage{caption}
\usepackage[numbers,sort&compress]{natbib}
\usepackage{tabularx}
\usepackage{hyperref}
\hypersetup{
	colorlinks=true,
	linkcolor=cyan,
	filecolor=blue,      
	urlcolor=red,
	citecolor=green,
}

\begin{document}

\def\xc{\textcolor{black}}
\newcommand{\red}[1]{{\color{black}{#1}}}

\definecolor{blue}{RGB}{68,114,196}
\definecolor{orange}{RGB}{237,125,49}
\def\etal{\textit{et al.}}
\def\etc{\textit{etc}}
\def\ie{\textit{i.e.}}
\def\eg{\textit{e.g.}}

\title{
{SITA: Structurally Imperceptible and Transferable \\ Adversarial Attacks for Stylized Image Generation}

\author{Jingdan~Kang,
        Haoxin~Yang,
        Yan Cai,
        Huaidong~Zhang,
        Xuemiao~Xu, \\
        Yong Du, 
        Shengfeng He,~\IEEEmembership{Senior Member,~IEEE}
        }
\thanks{\textit{Accepted by IEEE Transactions on Information Forensics and Security.}}
\thanks{The work is supported by National Natural Science Foundation of China (No.62302170), Guangdong Basic and Applied Basic Research Foundation (No.2024A1515010187), Guangzhou Basic and Applied Basic Research Foundation (No.2024A04J3750), Guangdong Provincial Natural Science Foundation for Outstanding Youth Team Project (No.2024B1515040010), China National Key R\&D Program (Grant No.2023YFE0202700, 2024YFB4709200), NSFC Key Project (No.U23A20391), Key-Area Research and Development Program of Guangzhou City (No.2023B01J0022), Guangdong Natural Science Funds for Distinguished Young Scholars (Grant 2023B1515020097), the AI Singapore Programme under the National Research Foundation Singapore (Grant AISG3-GV-2023-011), and the Lee Kong Chian Fellowships. \textit{Jingdan~Kang and Haoxin~Yang contributed equally to this work. Corresponding authors: Huaidong Zhang and Xuemiao Xu.}}
\thanks{Jingdan~Kang, Yan~Cai, and Huaidong~Zhang are with the School of Future Technology, South China University of Technology, Guangzhou, China. E-mail: jingdankang6@gmail.com; yancai3113@gmail.com; huaidongz@scut.edu.cn.}
\thanks{Haoxin~Yang and Xuemiao~Xu are with the School of Computer Science and Engineering, South China University of Technology, Guangzhou, China. E-mail: yhx1996@outlook.com; xuemx@scut.edu.cn.}
\thanks{Yong~Du is with the School of Computer Science and Technology, Ocean University of China, Qingdao, China. E-mail: csyongdu@ouc.edu.cn.
} 
\thanks{Shengfeng He is with the School of Computing and Information Systems, Singapore Management University, Singapore. E-mail: shengfenghe@smu.edu.sg.}
}

\markboth{IEEE TRANSACTIONS ON INFORMATION FORENSICS AND SECURITY}%
{Kang \MakeLowercase{\textit{et al.}}:SITA: Structurally Imperceptible and Transferable Adversarial Attacks for Stylized Generation}

\maketitle
\begin{abstract}
Image generation technology has brought significant advancements across various fields but has also raised concerns about data misuse and potential rights infringements, particularly with respect to creating visual artworks. Current methods aimed at safeguarding artworks often employ adversarial attacks. However, these methods face challenges such as poor transferability, high computational costs, and the introduction of noticeable noise, which compromises the aesthetic quality of the original artwork.
To address these limitations, we propose a \textbf{S}tructurally \textbf{I}mperceptible and \textbf{T}ransferable Adversarial (\textbf{SITA}) attacks. SITA leverages a CLIP-based destylization loss, which decouples and disrupts the robust style representation of the image. This disruption hinders style extraction during stylized image generation, thereby impairing the overall stylization process. Importantly, SITA eliminates the need for a surrogate diffusion model, leading to significantly reduced computational overhead. The method’s robust style feature disruption ensures high transferability across diverse models.
Moreover, SITA introduces perturbations by embedding noise within the imperceptible structural details of the image. This approach effectively protects against style extraction without compromising the visual quality of the artwork. 
Extensive experiments demonstrate that SITA offers superior protection for artworks against unauthorized use in stylized generation. It significantly outperforms existing methods in terms of transferability, computational efficiency, and noise imperceptibility. Code is available at \url{https://github.com/A-raniy-day/SITA}.


\end{abstract} 
\section{Introduction}

\IEEEPARstart{R}ecent advancements in diffusion models~\cite{sohl2015deep, ho2020denoising} have greatly enhanced visual content creation, particularly in areas such as content customization and stylized image generation. Techniques such as DreamBooth~\cite{ruiz2022dreambooth}, T2I-Adapter~\cite{mou2023t2i}, and Textual-Inversion~\cite{gal2022image} enable users to personalize generated images with ease, marking a significant leap in the efficiency and versatility of image creation.
However, these advancements have also raised concerns over data security and intellectual property rights, particularly among artists. Diffusion-based stylization techniques can easily replicate an artist's distinctive style, resulting in generated images that may infringe on their intellectual property.

\begin{figure}[t]
    \centering	
    \footnotesize{
    \begin{tabular}{p{.0cm}}
        \rotatebox{90}{\hspace{.5cm}Original\hspace{.5cm}} \\
        \rotatebox{90}{\hspace{.5cm}SITA\hspace{.5cm}} \\
    \end{tabular}
    }
    \subfloat[\scriptsize{Input}]{
     \begin{minipage}{0.22\linewidth}
     \includegraphics[width=\linewidth]{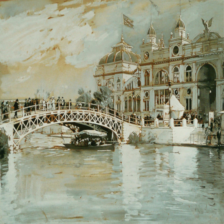}\\
     \vspace{-2mm}
     \includegraphics[width=\linewidth]{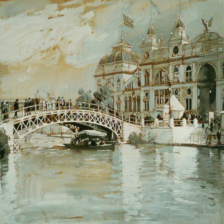}
     \end{minipage}
     }
    \hspace{-2mm}
    \subfloat[\scriptsize{Noise}]{\label{fig:threat_b}	
     \begin{minipage}{0.22\linewidth}
     \includegraphics[width=\linewidth]{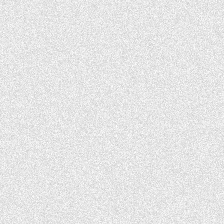}\\
     \vspace{-2mm}
     \includegraphics[width=\linewidth]{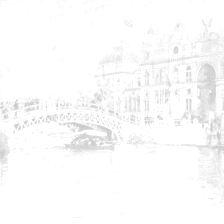}
     \end{minipage}
     }
    \hspace{-2mm}
    \subfloat[\scriptsize{Feature Map}]{\label{fig:threat_c}	
     \begin{minipage}{0.22\linewidth}
     \includegraphics[width=\linewidth]{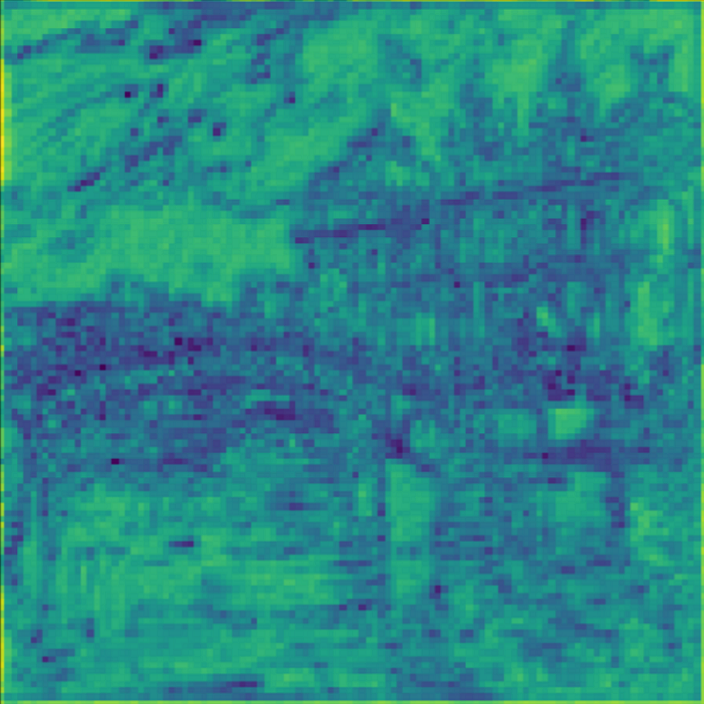}\\
     \vspace{-2mm}
     \includegraphics[width=\linewidth]{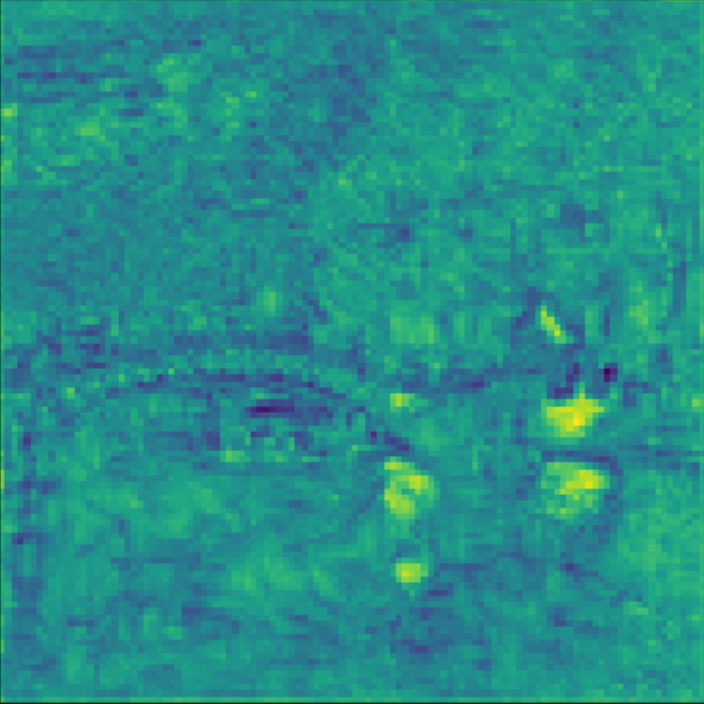}
     \end{minipage}
     }
    \hspace{-2mm}
    \subfloat[\scriptsize{Mimicry}]{\label{fig:threat_d}	
     \begin{minipage}{0.22\linewidth}
     \includegraphics[width=\linewidth]{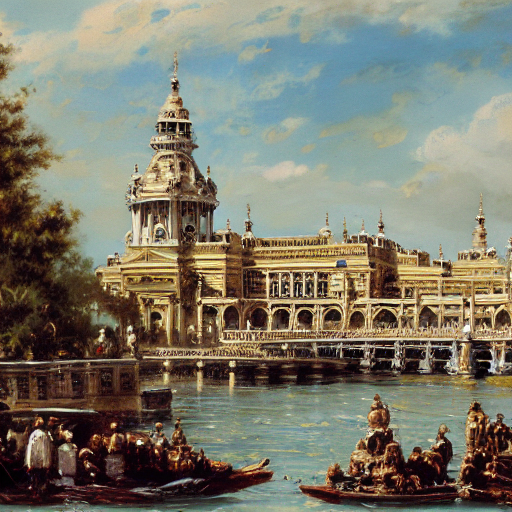}\\
     \vspace{-2mm}
     \includegraphics[width=\linewidth]{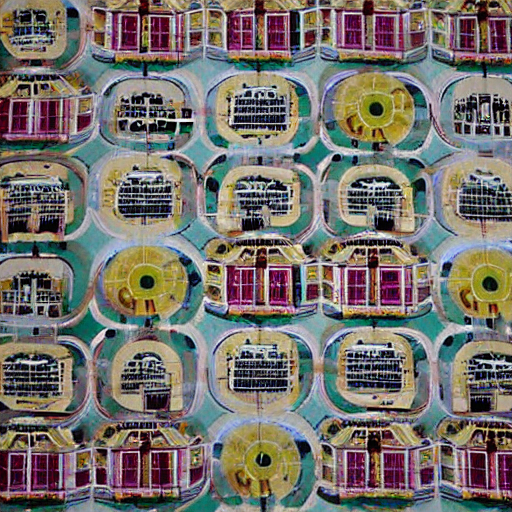}
     \end{minipage}
     }

    \caption{SITA is designed to interfere with the feature extraction process in image encoders. In (a), we present the original image overlaid with Gaussian noise alongside the sample generated using SITA. (b) depicts the noise map. (c) illustrates the feature map extracted by the latent diffusion model~\cite{rombach2022high}, clearly demonstrating how our introduced noise impedes the feature generation process. (d) showcases the style-imitated results derived from (a).}
    \label{fig:threat}
    
\end{figure}

To address concerns about unauthorized style imitation, recent methods have introduced adversarial noise~\cite{Szegedy2013IntriguingPO} to prevent models from copying an artist’s style. These approaches generally fall into two categories.

The first category includes evasion attacks~\cite{Goldblum2020DatasetSF}, which assume that the diffusion model’s parameters remain fixed during stylized image generation. 
For instance, AdvDM~\cite{liang2023adversarial} and SDS~\cite{xue2023toward} generate adversarial noise by applying gradient optimization to the diffusion model’s objective function, though they target different components of the diffusion models. In contrast, Glaze~\cite{shan2023glaze} and PID~\cite{li2024pid} protect artworks by modifying the distribution of the latent variables, employing distinct optimization objectives to achieve their respective goals.
While these methods are effective when model parameters remain fixed, they perform poorly when the stylized technique updates the model's parameters, such as in DreamBooth~\cite{ruiz2022dreambooth}.

The second category includes data poisoning attacks~\cite{Goldblum2020DatasetSF}, which target methods that involve fine-tuning for stylized image generation. Approaches like Anti-DreamBooth~\cite{van2023anti} and Metacloak~\cite{liu2024metacloak} use bi-level optimization strategies. They first minimize the diffusion model’s~\cite{rombach2022high} objective function and then maximize the generated error by adjusting the parameters. However, these methods are computationally expensive, requiring long training times that limit their practical application.
Additionally, these methods are ineffective against fixed-parameter approaches like T2I-Adapter~\cite{mou2023t2i}, which introduce an additional style adapter. In such poisoning attacks, adversarial examples depend on joint optimization with fine-tuned models. 

Despite recent advances, generating adversarial examples for stylized image generation faces three key challenges. First, designing a method specifically tailored for adversarial example generation insensitive to the version of the diffusion model is crucial for enhancing transferability, as the various diffusion-based stylized image generation models differ significantly in their parameters and architectures. Second, generating adversarial noise that remains imperceptible to the human eye while effectively preventing stylization models from replicating the original image’s style is a persistent challenge. Visible noise reduces the practical usability of images, particularly in artistic contexts. Finally, improving the computational efficiency of adversarial generation is essential for users with limited resources to protect their intellectual property, as existing methods are often computationally intensive and time-consuming, posing a barrier to widespread adoption.

To address these challenges, we propose SITA (\textbf{\textit{S}}tructurally \textbf{\textit{I}}mperceptible and \textbf{\textit{T}}ransferable \textbf{\textit{A}}dversarial attacks), a novel method for generating highly transferable adversarial examples across various diffusion-based stylized image generation techniques. SITA achieves imperceptible adversarial noise while maintaining high computational efficiency.
Our approach leverages the CLIP~\cite{radford2021learning} image encoder as a surrogate model, exploiting its powerful representational and style extraction capabilities~\cite{xu2023stylerdalle, chefer2022image, ramesh2022hierarchical, frans2022clipdraw}. We design a specialized loss function tailored for stylized image generation that decouples image style from content. This loss guides the generation of adversarial examples by maximizing the stylistic difference between the original reference image and the adversarial example within the CLIP representation space, leading to failure feature extraction in the latent diffusion model (see Fig.~\ref{fig:threat_c}). This not only enhances optimization efficiency but also improves the imperceptibility of the adversarial noise, ensuring that diffusion models cannot replicate the style of the reference image.
Since CLIP is trained on vast amounts of text-image alignment data and aligns well with text-to-image diffusion models, adversarial examples generated through CLIP representations transfer effectively across different diffusion-based stylization techniques.
To further minimize the perceptibility of adversarial noise, we introduce a structure perception loss, which confines the noise to regions of the image with pronounced color transitions and structural details, reducing the visibility of artifacts (see Fig.~\ref{fig:threat_b}). Crucially, SITA disrupts the CLIP image encoder without requiring extensive computations across the entire diffusion model, significantly improving both transferability and computational efficiency.
Extensive experiments demonstrate that SITA outperforms state-of-the-art techniques in terms of adversarial effectiveness, transferability, noise imperceptibility, and computational efficiency.

In summary, our contributions are fourfold:
\begin{itemize} 
\item We introduce SITA, a novel method that delivers imperceptible, transferable, and computationally efficient adversarial attacks. SITA effectively safeguards against copyright infringement in diffusion-based stylized image generation. 
\item We propose an implicitly decoupled destylization loss, which separates style and content features within the representation space of the CLIP image encoder. This approach disrupts the robust style representation of reference images, preventing various diffusion-based stylization methods from capturing and replicating the original style. 
\item We introduce a structure perception loss that confines adversarial noise to structural details, minimizing alterations in uniform regions and making the noise less perceptible to the human eye. 
\item Extensive experiments demonstrate the effectiveness of SITA across a wide range of stylized generation tasks. Our method successfully protects artworks from style extraction by generative models while maintaining high visual quality. 
\end{itemize}
\section{Related Work}
\subsection{Diffusion-based Image Generation}
Diffusion models~\cite{sohl2015deep, ho2020denoising, song2020denoising, rombach2022high} have recently garnered attention for their capacity to generate diverse and high-quality samples. The Denoising Diffusion Probabilistic Model (DDPM)~\cite{ho2020denoising} employs a Markov chain to gradually introduce Gaussian noise into images, subsequently learning to reverse this process for data reconstruction. The Denoising Diffusion Implicit Model (DDIM)~\cite{song2020denoising} streamlines this inverse process by optimizing and blending learning objectives, thus expediting inference. Furthermore, the Latent Diffusion Model (LDM)~\cite{rombach2022high} enhances efficiency by implementing diffusion in latent space, balancing computational demands with model quality and versatility.

\subsection{Personalized and Stylized Diffusion-based Generation}
Personalized and stylized diffusion-based generation aims to generate corresponding images based on given specific conditions. In stylized image generation, this often involves replicating the style of a specific artist.
Traditional approaches employ deep neural networks (DNNs) to extract patterns from a target style, which are then used as conditions by generative adversarial networks (GANs) to synthesize images in that style~\cite{deng2022stytr2,gatys2016image, zhang2022domain,xiao2022appearance}. 
Recent advancements in diffusion models have showcased significant generative capabilities, leading to two primary techniques for stylized generation. The first method involves fine-tuning the LDM with specific styles, exemplified by DreamBooth~\cite{ruiz2023dreambooth}, which refines pre-trained models using a small set of images to better replicate desired style features. Textual-Inversion~\cite{gal2022image} operates similarly, maintaining static LDM parameters while updating targeted ``pseudo-words" for stylization. The second method extends beyond the LDM framework, as seen in the T2I-Adapter~\cite{mou2023t2i} approach, which employs a conditional adapter to guide generation by encoding style images and leveraging a cross-attention mechanism. While these advanced conditional control generators facilitate the production of tailored visual content, their use also raises potential concerns about copyright infringement, particularly when applied to unauthorized data.
   
\subsection{Traditional Adversarial Attacks}
Adversarial attacks are engineered to mislead or manipulate machine learning models by introducing meticulously designed perturbations into the input data. These perturbations, though generally subtle and imperceptible to humans, can significantly disrupt model functionality, leading to errors or misclassifications~\cite{goodfellow2014explaining, madry2017towards, Carlini2016TowardsET, moosavi2016deepfool, papernot2016limitations, ma2023transferable, weng2023logit, chen2024diffusion}. Traditionally, perturbation magnitudes have been quantified using pixel-level norms. However, recent studies indicate that pixel-level norms may not accurately reflect human perceptual sensitivity~\cite{zhang2024perception, luo2022frequency, duan2021advdrop, long2022frequency}. In response, this paper introduces a Structure Perception Loss, which extends low-frequency constraints~\cite{luo2022frequency} to establish dual-domain regularization, effectively suppressing adversarial noise in both homogeneous regions and structural details. By aligning with the characteristics of the human visual system, this approach is designed to minimize the perceptual impact of perturbations, ensuring that the generated adversarial noise remains largely imperceptible to human observers.

\subsection{Adversarial Attacks on Diffusion-based Image Generation}
As visual content generation continues to advance, copyright concerns in visual data have become increasingly prominent. In response to unauthorized diffusion-based image generation, researchers have developed adversarial techniques, including evasion attacks and data poisoning~\cite{Goldblum2020DatasetSF}.

Evasion attacks, such as AdvDM~\cite{liang2023adversarial}, SDS~\cite{xue2023toward}, Glaze~\cite{shan2023glaze}, and PID~\cite{li2024pid} operate under the assumption that model parameters remain frozen. 
AdvDM~\cite{liang2023adversarial} pioneers the use of adversarial examples for diffusion models, maximizing the objective function to hinder accurate style extraction. 
SDS~\cite{xue2023toward} simplifies protection by employing score distillation sampling, bypassing gradient computations in denoising modules to reduce overhead. 
Glaze~\cite{shan2023glaze} merges style transfer with adversarial attack, minimizing feature discrepancies between original and stylized versions to mislead stylized generation models. 
PID~\cite{li2024pid}, on the other hand,  adjusts the mean and variance of the visual encoder’s latent distribution, offering prompt-independent protection.
However, these methods rely on the assumption of frozen model parameters, which is a limiting factor in practice. Their effectiveness decreases when models are fine-tuned (e.g., DreamBooth~\cite{ruiz2023dreambooth}), as parameter updates disrupt the precomputed adversarial patterns. 
.

On the other hand, data poisoning attacks, which assume that model parameters will change, include methods like Anti-DreamBooth~\cite{van2023anti} and Metacloak~\cite{liu2024metacloak}. These approaches use a bi-level optimization strategy: the inner optimization minimizes the Latent Diffusion Model's (LDM) objective to emulate parameter changes during the stylized generation process, while the outer optimization maximizes the generation error based on the adjusted parameters. Metacloak further improves the transferability of attacks by integrating multiple surrogate models during the inner optimization. However, these methods rely on complex bi-level optimization, with training sessions even lasting several hours, limiting their practical applications. Additionally, since both methods are designed for fine-tuning the original LDM architecture, their effectiveness is significantly reduced when applied to T2I-Adapter~\cite{mou2023t2i}, a technique that employs an additional encoder for style extraction while keeping the LDM parameters unchanged.

These methods present notable limitations. AdvDM, Glaze, and Anti-DreamBooth, as white-box attacks, require full access to the target model's parameters, which restricts their broader applicability. On the other hand, Metacloak, a black-box attack, demands multiple diffusion surrogate models, resulting in high computational costs. Additionally, none of these approaches have designed a specific style loss for stylized image generation, instead relying on the entire LDM, which demands substantial resources—posing difficulties for users with limited computational power. Furthermore, these methods primarily depend on basic metrics, such as the $L_{\infty}$ norm, to constrain noise size, potentially compromising the visual quality of the resulting artworks. They also lack thorough analyses of the visual impact of adversarial noise.

\section{Proposed Method}
\label{method}

\subsection{Preliminaries}
\subsubsection{Latent Diffusion Models}
The conditional generation process in latent diffusion models (LDMs)~\cite{rombach2022high} begins by sampling a latent variable $ z_T $ from a standard Gaussian distribution, $ z_T \sim \mathcal{N}(0, I) $. Given a conditioning input $ c $, the model refines this latent variable iteratively through a denoising process. At each timestep $ t $, the latent code is updated as follows:  

\begin{equation}  
z_{t-1} = \frac{1}{\sqrt{\alpha_t}} \left( z_t - \frac{1 - \alpha_t}{\sqrt{1 - \bar{\alpha}_t}} \epsilon_\theta (z_t, t, c) \right),  
\end{equation}  
where $ z_t $ represents the latent variable at timestep $ t $, and $ \alpha_t $ is a noise schedule parameter that controls the variance of the noise introduced during the forward diffusion process. The term $ \bar{\alpha}_t $ denotes the cumulative product of these noise schedule parameters, defined as $ \bar{\alpha}_t = \prod_{s=1}^{t} \alpha_s $. The function $ \epsilon_\theta $ represents the denoising model, which estimates and removes the noise while incorporating the conditioning information $ c $.  
As the iterative denoising process unfolds, the influence of the noise term gradually diminishes, allowing the latent variable to progressively refine and ultimately converge to $z_0$. Once the latent representation has stabilized, it is passed through the model’s variational autoencoder decoder to generate the final image, ensuring both faithful adherence to the conditioning input $c$ and the preservation of realistic visual quality.

\subsubsection{Problem Formulation}
In this paper, our primary objective is to inhibit a stylized image generation model from generating images that replicate the styles of reference images, while preserving as much of the visual effect to humans from the style references as possible. 
To achieve this objective, we first identified that the CLIP image encoder effectively extracts general and robust image features. Leveraging this capability, we employed the CLIP image encoder as our surrogate model and introduced a novel CLIP-based destylization loss to guide the generation of adversarial examples. Additionally, to further enhance the visual quality of these generated samples, we proposed a structure perception loss that aligns with human visual perception. This loss function constrains adversarial noise to detailed regions that are less perceptible to the human eye. 
This optimization-based strategy provides enhanced flexibility in controlling adversarial perturbations, allowing for more precise suppression of stylization effects while maintaining high visual fidelity.
As a result, our problem formulation can be distilled into the following definition:
\begin{equation}
\begin{aligned}
        X_{adv} :=  \mathop{\arg \min } \limits_{X_{adv}}\ & \{ \lambda \mathcal{L}_{destyle}(X_s,\ X_{adv}) + \\
                            & \mathcal{L}_{per}(X_s,\ X_{adv}) \},
\end{aligned}
\end{equation}
where $X_{adv}$ represents the adversarial example, $X_{s}$ denotes the original style reference image, $\mathcal{L}_{destyle}$ and $\mathcal{L}_{per}$ denote the novel CLIP-based destylization loss and structure perception loss introduced in this paper, and $\lambda$ is a coefficient employed to adjust $\mathcal{L}_{destyle}$.

\subsection{Overview}
\subsubsection{SITA}
\begin{figure*}[!t]
   \begin{center}
      \includegraphics[width=1.\linewidth]{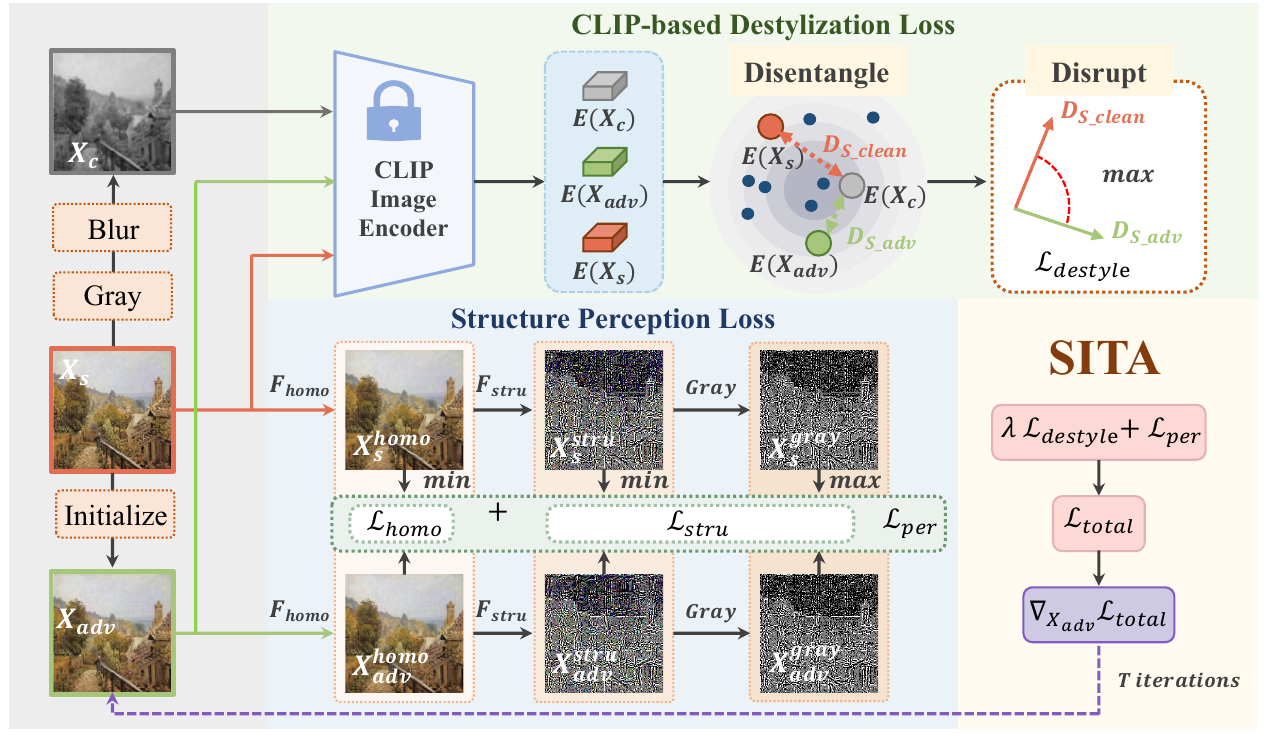}
   \end{center}
   \caption{Overview of the proposed SITA. We begin by initializing the adversarial style reference image $X_{adv}$ from the clean style reference image $X_s$ and extracting the content $X_c$ of $X_s$. Next, we perform style disentanglement between $X_s$, $X_c$, and $X_{adv}$ in the CLIP feature space, calculating the CLIP-based Destylization Loss to precisely manipulate style information.  Additionally, we decompose the regions of both $X_s$ and $X_{adv}$ into homogeneous areas and structural details and apply the Structure Perception Loss to constrain adversarial noise placement. This ensures the noise becomes less perceptible while maintaining the fidelity of $X_{adv}$. After $T$ iterations of optimization, we obtain an adversarial example that is both imperceptible in terms of noise and free of the original style information.}
   \label{fig:framework}
   \end{figure*}

To prevent stylized generation while preserving the visual appeal of artworks, we propose an efficient framework called the Structurally Imperceptible and Transferable Adversarial (SITA) attack. This framework generates imperceptible and transferable adversarial examples specifically designed for diffusion-based stylized generation.
As illustrated in Fig.~\ref{fig:framework}, our framework utilizes the CLIP image encoder as a surrogate model for generating adversarial examples. Given that CLIP is pre-trained on a large-scale text-image dataset, it has a general feature representation and is well-aligned with the task paradigm of text-to-image diffusion models. Consequently, we developed a CLIP-based destylization loss based on CLIP's feature representation. This loss effectively decouples style features from content features, guiding the generation of de-styled adversarial examples and significantly reducing the perceptibility of adversarial noise. The strong representational capabilities of CLIP enable the adversarial examples generated to demonstrate excellent generalization properties.
To further reduce the perceptibility of adversarial noise, we incorporate a sophisticated structure perception loss, which restricts noise primarily to subtle structural detail areas.

\subsubsection{Threat Model} 
We state the threat model in this paper according to the following assumptions:
\begin{itemize}
    \item The artists wish to publicly share their artworks while preserving their visual quality but also seek to prevent stylized image generation models from infringing upon their artistic style.
    \item The artists have access only to the weights and architecture of the CLIP image encoder and are unaware of the version of the diffusion model or specific stylized generation method that the infringer may be using.
\end{itemize}

\subsection{CLIP-based Destylization Loss}
\subsubsection{Design Principles}
Before presenting the details of our destylization loss, we first outline the key observations informing our design.

\textbf{Guiding Question 1: Why do we aim to disentangle and disrupt style features?} Previous methods \cite{madry2017towards, Carlini2016TowardsET, goodfellow2014explaining} for generating adversarial examples often base on classification models. In the case of classification models, the output logit is already disentangled, meaning that the logit corresponding to each class is independent. Consequently, guiding the generation of adversarial examples is straightforward by simply altering the logit.
However, the features used for stylized image generation are interdependent and encompass all features of the image. Blindly manipulating these features (for instance, by maximizing the distance between them) can result in generated adversarial examples that are noisy and difficult to converge \cite{zhou2023downstreamagnostic, MoosaviDezfooli2016UniversalAP, van2023anti, shan2023glaze, liang2023adversarial, liu2024metacloak}, due to the absence of effective loss guidance and feature decoupling methods. Our objective is not only to deter the learning of style features but also to ensure the visual quality of the artwork. To achieve this, we must first disentangle the style features and perform more precise manipulations.

\textbf{Guiding Question 2: Why do we choose CLIP image encoder as the surrogate model?} 
CLIP is trained on large-scale text-image alignment datasets and offers excellent generalization and representation capabilities. Previous works have demonstrated it can effectively extract style information from images~\cite{patashnik2021styleclip,frans2022clipdraw,kwon2022clipstyler,vinker2022clipasso}. With our destylization loss, we can successfully extract, disentangle, and disrupt style information.
Second, considering that the text-to-image latent diffusion model (LDM) operates within a text-to-image task paradigm and that CLIP is trained on text-image alignment, it can effectively model the relationship between these two modalities. Consequently, adversarial examples generated using CLIP as a surrogate model are well-suited for application in the diffusion stylized generation task.
Finally, since different LDM versions use the CLIP text encoder for conditional injection, adversarial examples generated by the CLIP image encoder show strong generalization across various stylized generation methods and LDM versions, as supported by our experiments.

\subsubsection{Objective Function}
In the field of stylized image generation, an image typically consists of two complementary components: content and style \cite{Gatys2015ANA, Johnson2016PerceptualLF, Huang2017ArbitraryST}. While identifying the content of an image is straightforward \cite{Wang2023StyleDiffusionCD}, characterizing the style of an image poses a complex challenge. This complexity arises from the fact that different works of art exhibit distinct styles, making explicit definition a difficult task.
To provide an effective destylization loss for the samples generated by SITA, we propose a loss function that implicitly captures the distance between image style and its content within the representation space of the CLIP image encoder.
Note that color and brushwork are critical elements of artistic style \cite{Meyer1979-MEYTAT-4}, and numerous style classification tasks consider these two features as pivotal criteria \cite{Deac2006FeatureSF, Zujovic2009ClassifyingPB, Agarwal2015GenreAS}. Consequently, the effacement of these key elements can facilitate the approximate extraction of content.
Thus, we explicitly derive the content of the style reference image $X_c$ by converting it to grayscale and applying a Gaussian blur filter to eliminate the color and brushstroke details as follows:

\begin{equation}
X_c = \mathrm{Blur}(\mathrm{Gray}(X_s)),
\label{eq:content}
\end{equation}
where $\mathrm{Gray}(\cdot)$ denotes the operation of converting the image to grayscale, and $\mathrm{Blur}(\cdot)$ refers to the Gaussian blur filter.
This approach circumvents the introduction of excessive additional learning parameters, which is crucial for maintaining the framework's computational efficiency.

Next, the representation space of the CLIP image encoder can serve as an effective metric space for implicitly measuring the ``style distance'' between the original image and its content. This style distance can be represented as:
\begin{equation}
\mathcal{D}_{S\_clean}(X_s, X_c) = E_{clip}(X_s) - E_{clip}(X_c),
\label{eq:sd_clean}
\end{equation}
where $E_{clip}$ is the CLIP image encoder.

Similarly, the style distance between the adversarial image $X_{adv}$ and the content image $X_c$ of $X_s$ can be expressed as:
\begin{equation}
\mathcal{D}_{S\_adv}(X_{adv}, X_c) = E_{clip}(X_{adv}) - E_{clip}(X_c).
\label{eq:sd_adv}
\end{equation}
With these two style distances, we define our destylization loss $\mathcal{L}_{destyle}$ as:
\begin{equation}
\mathcal{L}_{destyle} = \mathrm{Cos} ({D}_{S\_adv}(X_{adv}, X_c), \mathcal{D}_{S\_clean}(X_s, X_c) ),
\label{eq:destylization}
\end{equation}
Here, $\mathrm{Cos}(\cdot)$ denotes the cosine similarity metric and it is suitable for CLIP's high-dimensional representation space. By utilizing cosine similarity, we can more precisely control the similarity between two style distances: the distance between the adversarial example and the image content, and the distance between the style reference image and the image content. This approach guides the generated adversarial sample to deviate from the original style in the feature space, thus achieving the de-style adversarial example.

\subsection{Structure Perception Loss}

Preserving the visual impact of the protected artworks in comparison to the original style reference images is a crucial consideration. To address this, we introduce a structure perception loss to control the perceptibility of the generated adversarial noise and enhance the visual quality of the resulting samples.

\begin{figure}[t]
    \centering	
    \subfloat[]{\label{fig:hvs_a}	
     \begin{minipage}{0.25\linewidth}
     \includegraphics[width=\linewidth]{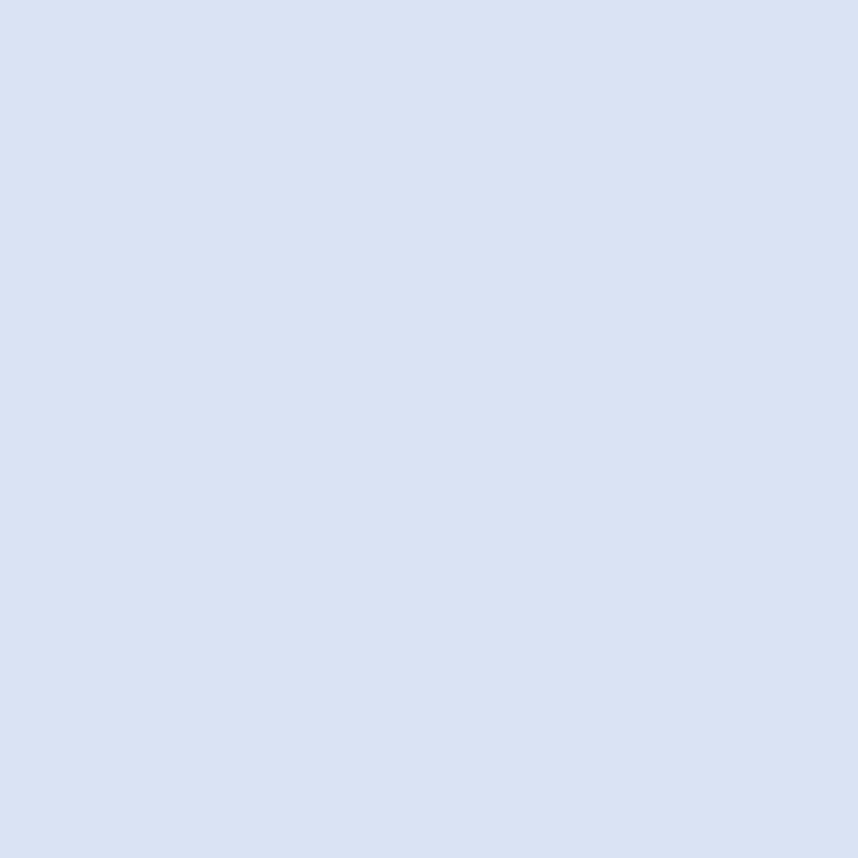}\\
     \vspace{-2mm}
     \includegraphics[width=\linewidth]{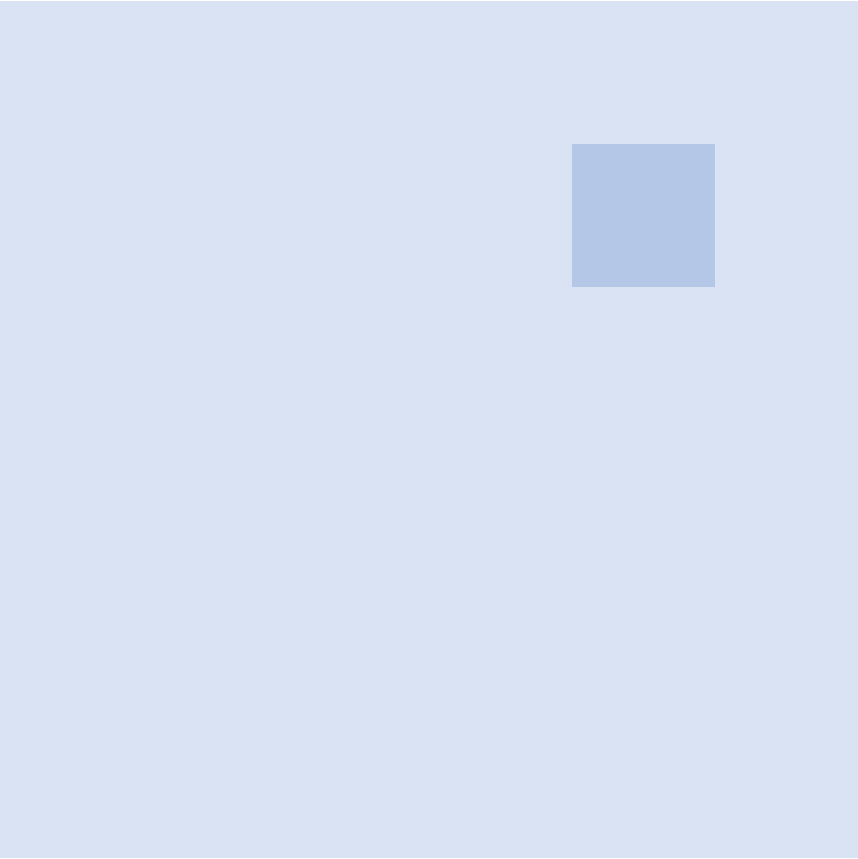}
     \end{minipage}
     }
    \subfloat[]{\label{fig:hvs_b}	
     \begin{minipage}{0.25\linewidth}
     \includegraphics[width=\linewidth]{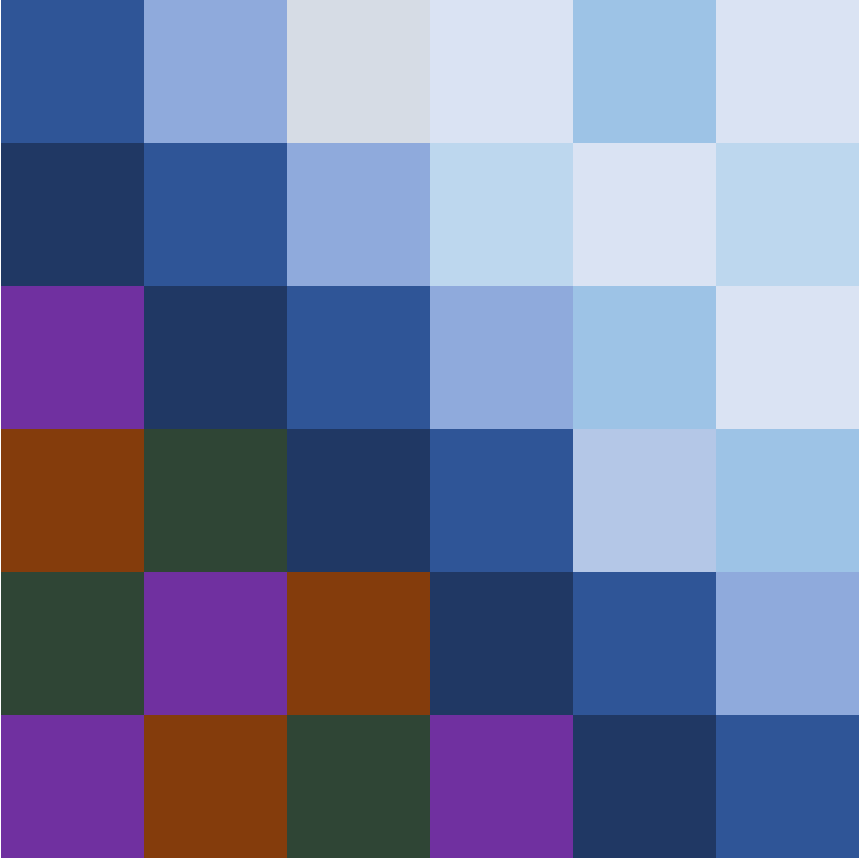}\\
     \vspace{-2mm}
     \includegraphics[width=\linewidth]{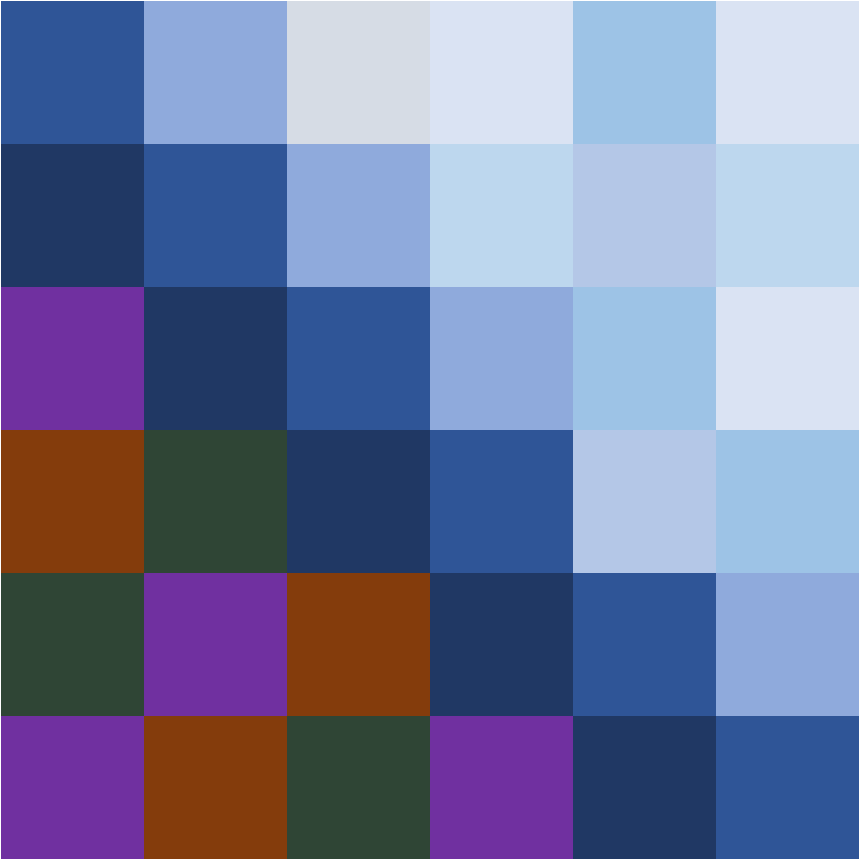}
     \end{minipage}
     }
    \subfloat[]{\label{fig:hvs_c}	
     \begin{minipage}{0.25\linewidth}
     \includegraphics[width=\linewidth]{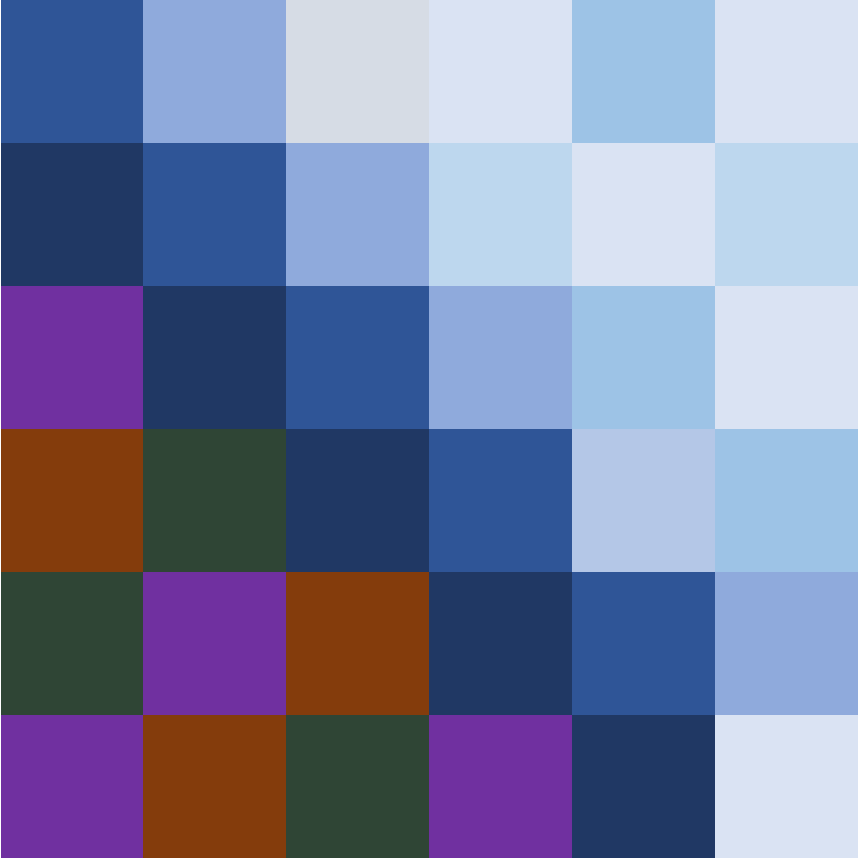}\\
     \vspace{-2mm}
     \includegraphics[width=\linewidth]{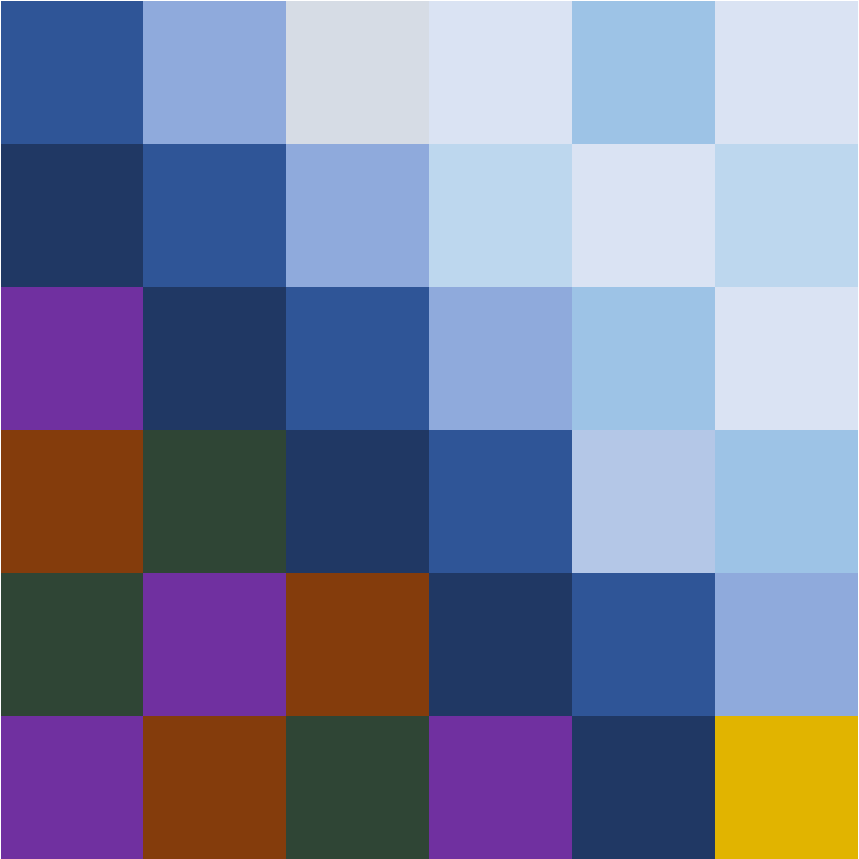}
     \end{minipage}
     }
    \caption{Diagram illustrating the impact of various types of noise on the human visual system across different regions.
    (a) depicts the visual impact of altering a single pixel within the homogeneous regions.
    (b) illustrates the visual consequences of identical pixel alterations occurring at the corresponding locations in the first column of the structural details.
    (c) portrays the visual effects resulting from changes in pixels of the same hue, as well as changes in pixels of different hues, while maintaining the same absolute difference pixel value as the original pixel.}
    \label{fig:hvs}	
\end{figure}

To begin, we delineate two distinct areas within the image: the homogeneous regions, where color changes occur more gradually,  and the structural details, characterized by sharp color transitions.
As illustrated in Fig.~\ref{fig:hvs_a} and Fig.~\ref{fig:hvs_b}, the human eye is particularly sensitive to subtle noise in large, uniform color regions (homogeneous regions), whereas it is challenging to discern color variations in regions with pronounced color shifts (structural details).
Consequently, we propose a structure perception loss to incorporate adversarial noise into the structural details to the greatest extent possible, thereby increasing the difficulty for the human eye to detect it.

To achieve more precise delineations of homogeneous regions and structural details, we first follow the approach outlined in \cite{luo2022frequency}. Specifically, we begin by extracting the low-frequency component of an input image $x$ using Discrete Wavelet Transform (DWT) \cite{shensa1992discrete}. This low-frequency component is then reconstructed back into the image domain through the Inverse Discrete Wavelet Transform (iDWT) \cite{shensa1992discrete}, yielding the homogeneous regions, denoted as:
\begin{equation}
F_{homo}(x) = \mathrm{iDWT}\left(\mathrm{DWT}(x)_{ll}\right),
\label{eq:homo}
\end{equation}
where $\mathrm{DWT}(\cdot)$ and $\mathrm{iDWT}(\cdot)$ refer to the operations of DWT and iDWT, respectively, and $ll$ signifies the low-frequency component.
Following~\cite{luo2022frequency}, the loss for the homogeneous regions between the style image $X_s$ and the adversarial image $X_{adv}$ can be formulated as:
\begin{equation}
\mathcal{L}_{homo} = \| F_{homo}(X_{adv}) - F_{homo}(X_s) \|_1.
\label{eq:homoloss}
\end{equation}
By optimizing this loss, the generated adversarial example remains consistent with the original image in the homogeneous regions, reducing the perceptibility of adversarial noise. 

As illustrated in Fig.~\ref{fig:hvs}, while the homogeneous constraint effectively introduces low-frequency restrictions to reduce the perceptibility of adversarial noise in homogeneous regions, it does not account for the structural details, which are also critical for the visual quality of adversarial examples. To address this, we propose an additional structural constraint to further enhance noise imperceptibility.
Specifically, as depicted in Fig.~\ref{fig:hvs_c}, although the absolute magnitudes of the variations between the two modified pixels might be identical, the perceptual impact on human vision is substantially diminished when the pixel's alteration results in a color transition within the same hue family. To address this, we first define how to compute the structural details of an input image $x$:

\begin{equation}
F_{stru}(x) = x - F_{homo}(x),
\label{eq:stru}
\end{equation}
where $F_{homo}(\cdot)$ corresponds to the operation defined in Eq.~\eqref{eq:homo}.

Subsequently, we introduce the structural details loss between $X_{adv}$ and $X_s$ as follow:
\begin{equation}
    \begin{aligned}
        \mathcal{L}_{stru} = & \| F_{stru}(X_{adv}) - F_{stru}(X_s) \|_1 \\
        & - \| \mathrm{Gray}(F_{stru}(X_{adv})) - \mathrm{Gray}(F_{stru}(X_{s})) \|_1,
    \end{aligned}
\label{eq:struloss}
\end{equation}
where $F_{stru}(\cdot)$ represents the operation defined in Eq.~\eqref{eq:stru}.
By optimizing this loss, the original hue is maintained while only the brightness of the color is changed, making it more difficult for the human eye to detect.

Thus, the structure perception loss can be denoted by:
\begin{equation}
    \mathcal{L}_{per} = \mathcal{L}_{homo} + \mathcal{L}_{stru}.
\label{eq:perloss}
\end{equation}

Our dual-domain formulation simultaneously enforces the imperceptibility of noise in homogeneous regions and preserves structural details, strategically aligning with the characteristics of the human visual system, thereby achieving enhanced adversarial stealth.

\renewcommand{\algorithmicrequire}{\textbf{Input:}}  
\renewcommand{\algorithmicensure}{\textbf{Output:}} 

\begin{algorithm}[t]
    \caption{Optimization Process of SITA.}
    \label{algo}
    \begin{algorithmic}[1]
    \Require Clean style reference image $X_s$, iteration steps $T$, coefficient $\lambda$, CLIP image encoder ${E}_{clip}$.
    \Ensure Adversarial style reference image $X_{adv}$.
    \State Initialize $X_{adv}^0 \leftarrow X_{S}$.
    \State Calculate the content $X_c$ of $X_s$ by Eq.~\eqref{eq:content}.
    \State Calculate the style distance $\mathcal{D}_{S\_clean}$ between $X_s$ and $X_c$ by Eq.~\eqref{eq:sd_clean}.
    \For{$t = 0$ \textbf{to} $T$}
        \State Calculate the style distance $\mathcal{D}_{S\_adv}$ between $X_{adv}^t$ and $X_c$ by Eq.~\eqref{eq:sd_adv}.
        \State Calculate the CLIP-based Destylization Loss $\mathcal{L}_{destyle}$ by Eq.~\eqref{eq:destylization}.
        \State Calculate the Structure Perception Loss $\mathcal{L}_{per}$ by Eq. ~\eqref{eq:homo} - Eq.~\eqref{eq:perloss}.
        \State Optimize $X_{adv}^t$ by minimizing $\mathcal{L}_{total}$ in Eq.~\eqref{eq:total}. 
    \EndFor 
    \State return $X_{adv}^T$.
    \end{algorithmic}
    \end{algorithm}

\subsection{Optimization Process of SITA}
\label{obj_function}
The objective function of our proposed SITA, designed against diffusion-based stylized image generation can be formulated as:  
\begin{equation}
    \mathcal{L}_{total} = \lambda\mathcal{L}_{destyle} + \mathcal{L}_{per},
    \label{eq:total}
\end{equation}
where $\lambda$ controls the trade-off between the decoupling destylization loss $\mathcal{L}_{destyle}$ and the structure perception loss $\mathcal{L}_{per}$.  

To generate an adversarial example $X_{adv}$ that effectively resists diffusion-based style generation while preserving the visual similarity to the original style reference image $X_s$, we employ an iterative optimization process that minimizes $\mathcal{L}_{total}$ to generate $X_{adv}$. Unlike explicit-constraint attack methods~\cite{goodfellow2014explaining, madry2017towards, liang2023adversarial, li2024pid, xue2023toward, van2023anti, liu2024metacloak}, which impose a fixed perturbation budget, SITA integrates perturbation control directly into the optimization objective. Rather than relying on rigid bounds, our approach dynamically adapts the noise magnitude, ensuring that the perturbation is minimized to the smallest necessary level. This dynamic control significantly reduces the perceptual distortions typically introduced by adversarial modifications. A detailed description of the SITA optimization process is outlined in Algorithm~\ref{algo}.

\section{Experiments}

\subsection{Implementation Details}
\noindent \textbf{Dataset.} 
Following by previous works~\cite{shan2023glaze, liang2023adversarial}, we select the WikiArt \cite{saleh2015large} dataset, known for its extensive collection of publicly available artistic works, ensuring close alignment with the context of our study. Additionally, we include another art dataset, ArtBench~\cite{liao2022artbench}, to demonstrate the generalizability of our approach. All samples are resized to 224$\times$224.

\noindent \textbf{Models.} 
We select three representative diffusion-based stylized generation methods for evaluation: fine-tuning the diffusion model with DreamBooth~\cite{ruiz2023dreambooth}, inverting the style into textual space using Textual-Inversion~\cite{gal2022image}, and pre-training a style encoder using the T2I-Adapter~\cite{mou2023t2i}.
For Textual-Inversion and DreamBooth, we randomly chose 200 sets of images from WikiArt. Following the Textual-Inversion~\cite{gal2022image}, we fine-tune a pseudo-word $p$ for each image set over 5,000 iterations to capture the unique style of the artist's paintings. With the prompt ``a painting in the style of $p$", we generated 50 images per pseudo-word, yielding a total of 10,000 generated images per method for our metrics.
For DreamBooth, we fine-tune the model for 1,000 iterations per image set, with a learning rate of $5 \times 10^{-7}$ and a batch size of 2, using the prompt ``a painting in the style of sks." Each fine-tuned model generated 50 images, resulting in 10,000 images for our metrics.
For the T2I-Adapter, we randomly select 10,000 images from WikiArt as style references and generate corresponding style transfer images for our metrics. 

Importantly, across all three stylized generation methods, the adversarial examples are generated using only a pre-trained, publicly available CLIP image encoder~\cite{radford2021learning}. Our method operates in a black-box setting, meaning it does not rely on internal access to the diffusion model itself.

\noindent \textbf{Experimental Settings.} As discussed in Section~\ref{obj_function}, our method eschews the conventional explicit perturbation budget, opting instead to integrate noise constraints directly into the objective function via the Structure Perception Loss. Specifically, we implicitly control the intensity of the noise by setting the learning rate and iteration steps. In our setup, we set iteration steps $T$ to 50 and set $\lambda$ to 100, with the Adam optimizer configured at a learning rate of 0.005. Please refer to the ablation study for the selection of hyperparameters. All experiments are conducted on NVIDIA A40 GPU using the PyTorch framework \cite{paszke2019pytorch}. 

\noindent \textbf{Baselines.}
In the realm of diffusion-based stylized image generation, we select two categories of state-of-the-art methods for comparison: AdvDM~\cite{liang2023adversarial}, Glaze~\cite{shan2023glaze}, PID~\cite{li2024pid}, and SDS~\cite{xue2023toward} for evasion attacks, and Anti-Dreambooth~\cite{van2023anti} and Metacloak~\cite{liu2024metacloak} for data poisoning attacks. For all implementations, we adhere to the hyperparameters settings provided in the official code. 

\subsection{Comparison with State-of-the-Art Methods}
\begin{figure*}[!t]
  \centering
  \includegraphics[width=\linewidth]{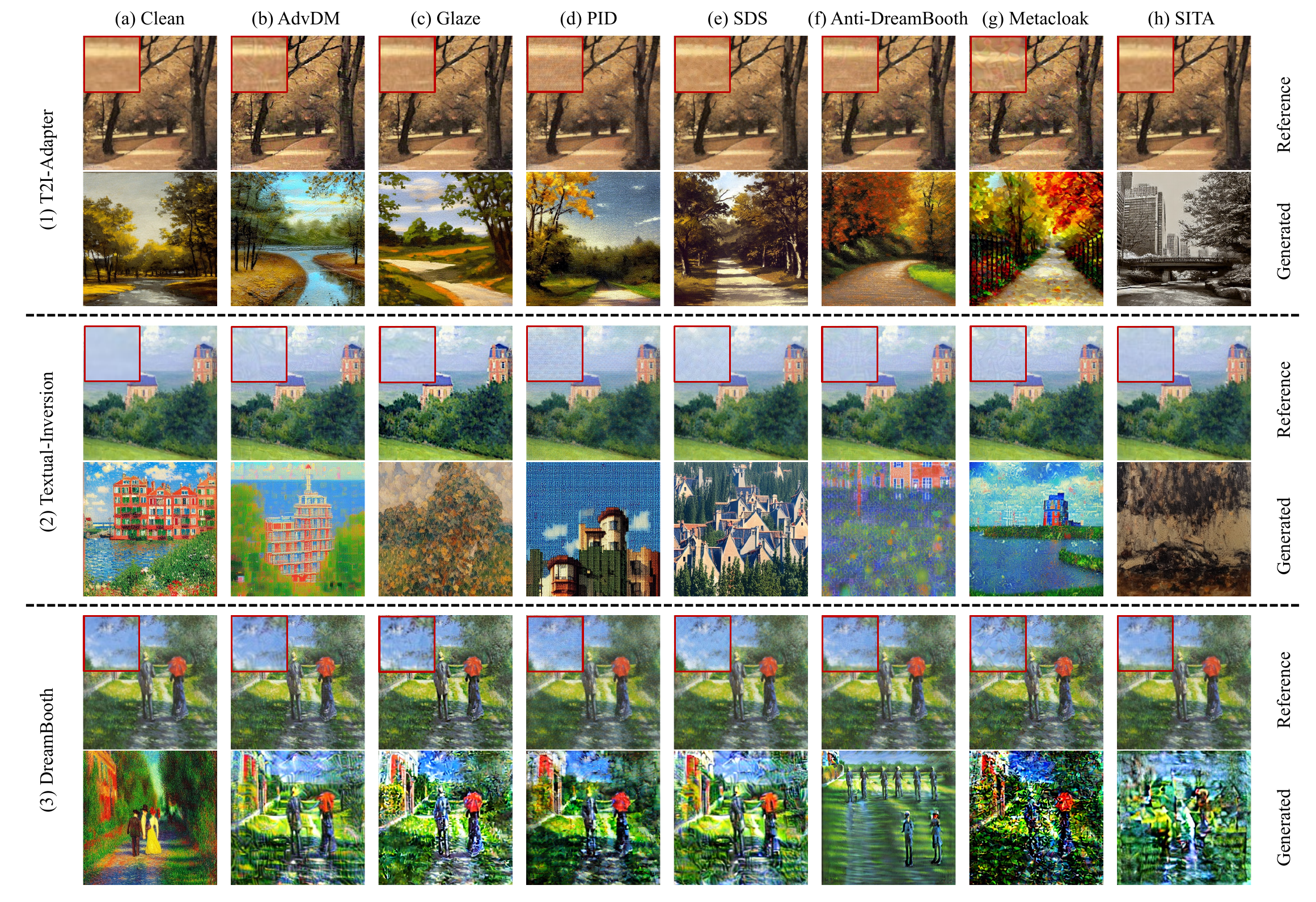} 
  \caption{ Visual results of different methods on the task of T2I-adapter~\cite{mou2023t2i}, Textual-Inversion~\cite{gal2022image} and DreamBooth~\cite{ruiz2022dreambooth}, based on various style reference images. From (a) to (h), the images represent the original style reference and its result, followed by the adversarial style references and outputs of AdvDM~\cite{liang2023adversarial}, Glaze~\cite{shan2023glaze}, PID~\cite{li2024pid}, SDS~\cite{xue2023toward}, Anti-DreamBooth~\cite{van2023anti}, Metacloak~\cite{liu2024metacloak}, and SITA, respectively.}
  \label{fig:comparison} 
\end{figure*}

\noindent\textbf{Qualitative Comparison.} In Fig.~\ref{fig:comparison}, we present the visual results for T2I-Adapter\cite{mou2023t2i}, Textual-Inversion~\cite{gal2022image}, and DreamBooth~\cite{ruiz2023dreambooth} on the WikiArt~\cite{saleh2015large} dataset. A closer examination of noise perceptibility reveals that the adversarial noise generated by AdvDM~\cite{liang2023adversarial}, PID~\cite{li2024pid}, Anti-DreamBooth~\cite{van2023anti}, and Metacloak~\cite{liu2024metacloak} is clearly noticeable, leading to a significant degradation in the visual quality of the artwork. While the noise introduced by Glaze~\cite{shan2023glaze} is more subtle due to its LPIPS constraint rather than a norm constraint, it still detracts from the overall image quality. Additionally, SDS~\cite{xue2023toward}, through its gradient descent strategy, somewhat improves the imperceptibility of the noise but still causes a smearing and blurring effect that is visible on the image.
In contrast, the results from our SITA method closely resemble the clean images, demonstrating the effectiveness of our structure perception loss in restricting noise to imperceptible structural details and minimizing its presence in uniform regions.

In terms of adversarial effectiveness, the results vary across tasks. Firstly, in the T2I-Adapter task, all protective methods except SITA fail, as the generated samples still have their styles extracted and replicated. This is primarily because these methods were developed based on the original LDM architecture, whereas T2I-Adapter uses an additional style adapter for style extraction, rendering the adversarial noise from other methods ineffective. Our method, however, disrupts the robust style information within the image itself, independent of the LDM architecture, preventing style replication more effectively.

Secondly, for the Textual-Inversion task, stylized results from all other adversarial examples struggle to entirely prevent the model from capturing the style of the reference images. Notably, our SITA approach demonstrates the strongest adversarial effect, completely impeding style replication, while the other methods still retain some stylistic traces of the original image. This can be attributed to our destylization loss, which decouples and disrupts style information more effectively, blocking style extraction at the core.

Finally, in the DreamBooth task, we observe that AdvDM and Glaze, which are based on evasion attacks, show diminished effectiveness as model parameters are updated. In contrast, the other methods, including ours, exhibit stronger adversarial performance in handling model parameter changes. Our method consistently demonstrates superior adversarial effectiveness across different customization tasks, highlighting its transferability. This underscores how our CLIP-based destylization loss effectively disrupts the robust style information embedded in images, ensuring stronger protection against style replication.

\begin{table*}[h]
\centering
    \caption{Comparison of the image similarity and perturbation norm among adversarial examples generated by AdvDM \cite{liang2023adversarial}, Glaze \cite{shan2023glaze}, PID~\cite{li2024pid}, SDS~\cite{xue2023toward}, Anti-DreamBooth\cite{van2023anti}, Metacloak\cite{liu2024metacloak} and our proposed SITA with the original style reference image, along with their corresponding computational costs. Bold font signifies the optimal results, underscore indicates suboptimal results.}
    \begin{tabular}{lrrrrrrr}
    \toprule
    Method          & SSIM$\uparrow$ & PSNR$\uparrow$  & MAE$\downarrow$  & Norm($L_2$)$\downarrow$     & Norm($L_\infty$)$\downarrow$      & Time(s)$\downarrow$       & GPU(GB)$\downarrow$      \\ \midrule
    AdvDM\cite{liang2023adversarial}           & 0.922          & 33.844          & 0.016            & 6.066                       & 0.031                             & 78                        & 19.8         \\
    Glaze\cite{shan2023glaze}           & {\ul 0.952}    & {\ul 36.874}    & {\ul 0.011}      & {\ul 4.953}                 & 0.046                             &  65                  & 5.90    \\
    PID\cite{li2024pid}              & 0.645    & 28.083    & 0.037      & 34.982                 & 0.050                             & 202                  & \textbf{0.45}    \\
    SDS\cite{xue2023toward}              & 0.931    & 34.634    & 0.015      & 5.832                 & 0.031                            &{\ul 28}                 & 10.7   \\
    Anti-DreamBooth\cite{van2023anti} & 0.874          & 34.604          & 0.017            & 16.505                      & \textbf{0.023}                    & 253                       & 45.55         \\
    Metacloak\cite{liu2024metacloak}       & 0.827          & 31.460          & 0.024            & 24.510                      & {\ul 0.043}                       & 6,333                      & 16.11         \\
    Ours            & \textbf{0.972} & \textbf{40.391} & \textbf{0.007}   & \textbf{4.236}              & 0.060                             & \textbf{24}               & {\ul 5.56} \\  \bottomrule
    \end{tabular}
\label{tab:image_similarity}
\end{table*}
\begin{table*}[]
\caption{Quantification results of FID \cite{heusel2017gans} and Image Alignment (IA) \cite{kumari2022multi} for image generation quality on WikiArt~\cite{saleh2015large} dataset and ArtBench~\cite{liao2022artbench} dataset.}
\begin{tabular}{l|cccccc|cccccc}
\toprule
                              & \multicolumn{6}{c|}{WikiArt \cite{saleh2015large}}                                                                                                                                                        & \multicolumn{6}{c}{ArtBench \cite{liao2022artbench}}                                                                                                                                                     \\ \cmidrule(l){2-13}  
\multirow{2}{*}{Method}                            & \multicolumn{2}{c}{T2I-Adapter \cite{mou2023t2i}} & \multicolumn{2}{c}{Textual-Inversion \cite{gal2022image}} & \multicolumn{2}{c|}{DreamBooth \cite{ruiz2022dreambooth}} & \multicolumn{2}{c}{T2I-Adapter \cite{mou2023t2i}} & \multicolumn{2}{c}{Textual-Inversion \cite{gal2022image}} & \multicolumn{2}{c}{DreamBooth \cite{ruiz2022dreambooth}} \\ \cmidrule(l){2-13} 
                                                   & FID$\uparrow$                    & IA$\downarrow$                 & FID$\uparrow$                        & IA$\downarrow$                     & FID$\uparrow$                        & IA$\downarrow$                     & FID$\uparrow$                    & IA$\downarrow$                 & FID$\uparrow$                        & IA$\downarrow$                     & FID$\uparrow$                       & IA$\downarrow$                     \\ \cmidrule(l){1-13} 
Clean                                              & 71.24                            & 0.890                          & 291.04                               & 0.746                              & 218.44                               & 0.745                              & 81.64                            & 0.902                          & 291.73                               & 0.787                              & 233.98                              & 0.781                              \\
AdvDM \cite{liang2023adversarial}  & 107.84                           & 0.809                          & 358.13                               & 0.725                              & 331.08                               & 0.732                              & 108.02                           & 0.810                          & 349.09                               & 0.743                              & 328.74                              & 0.778                              \\
Glaze \cite{shan2023glaze}         & {\ul 126.46}                     & {\ul 0.704}                    & 347.02                               & 0.691                              & 309.22                               & 0.734                              & {\ul 125.54}                     & {\ul 0.697}                    & 340.61                               & 0.704                              & 298.03                              & 0.763                              \\
PID \cite{li2024pid}               & 104.63                           & 0.843                          & 338.61                               & {\ul 0.683}                              & 294.70                               & 0.738                              & 103.28                           & 0.865                          & 337.20                               & 0.694                              & 287.84                              & 0.771                              \\
SDS \cite{xue2023toward}           & 100.55                           & 0.722                          & 332.68                               & 0.716                              & 334.38                               & 0.724                              & 102.80                           & 0.727                          & 342.68                               & 0.732                              & 316.79                              & 0.724                              \\
Anti-DreamBooth \cite{van2023anti} & 111.56                           & 0.752                          & {\ul 359.09}                         & 0.684                        & \textbf{385.83}                      & {\ul 0.701}                        & 113.24                           & 0.743                          & \textbf{361.92}                      & {\ul 0.641}                        & \textbf{348.47}                     & 0.739                              \\
Metacloak \cite{liu2024metacloak}  & 112.26                           & 0.721                          & 354.40                               & 0.687                              & 353.54                               & 0.709                              & 111.91                           & 0.726                          & 350.98                               & 0.706                              & 334.48                              & {\ul 0.722}                        \\
Ours                          & \textbf{141.36}                  & \textbf{0.672}                 & \textbf{364.54}                      & \textbf{0.679}                     & {\ul 357.43}                         & \textbf{0.691}                     & \textbf{141.89}                  & \textbf{0.676}                 & {\ul 351.62}                         & \textbf{0.627}                     & {\ul 336.37}                        & \textbf{0.701}                     \\ \bottomrule
\end{tabular}
\label{tab:fid}
\end{table*}

\noindent\textbf{Quantitative Comparison.}
We conduct comprehensive quantitative experiments to evaluate the performance of the SITA method and other approaches in terms of noise perceptibility, adversarial effectiveness, and computational efficiency. 

In the noise perceptibility tests, we use several image similarity metrics, including SSIM \cite{wang2004image}, PSNR \cite{gonzalez2008digital}, MAE, $L_2$ norm, and $L_{\infty}$ norm. The results in Table~\ref{tab:image_similarity} show that, except for the $L_{\infty}$ norm, our method achieves the best performance across all metrics, with Glaze ranking second. This is because we employed a noise perception constraint more aligned with human vision, rather than the $L_{\infty}$ noise constraint. Both SIDA and Glaze also demonstrated strong visual performance in previous qualitative experiments. Conversely, while Anti-DreamBooth performs best in the $L_{\infty}$ norm, followed by Metacloak, the noise they generated was highly noticeable in the visualized images. This suggests that the $L_{\infty}$ norm does not fully align with human visual perception as a noise constraint.
Instead of using the $L_{\infty}$ norm to constrain noise, we applied a structure perception loss that guides noise placement along edges and structural details, making it nearly imperceptible. This explains our method's superior performance in SSIM, PSNR, and MAE metrics.

In terms of computational efficiency, we evaluate the time and GPU memory required for each method to generate adversarial examples. The results in Table~\ref{tab:image_similarity} demonstrate that our SITA method is the most efficient, taking just 24 seconds to generate a single adversarial example while utilizing only 5.56GB of GPU memory. In comparison, although PID consumes only 0.45GB of GPU memory, its optimization time is approximately ten times longer than that of SITA. On the other hand, the SDS method enhances the gradient optimization strategy, generating an adversarial example in 28 seconds, but it requires 10.7GB of GPU memory. While both SDS, PID, and our SITA approach are relatively lightweight, with resource demands acceptable for most users, SITA strikes the best balance between image fidelity, adversarial effectiveness, and computational efficiency. In contrast, methods like Anti-Dreambooth and Metacloak, which employ bi-level optimization, significantly increase computational time. Notably, Anti-Dreambooth requires up to 45GB of GPU memory, making it impractical for most users due to the need for specialized hardware. While Metacloak requires less GPU memory, it takes several hours to generate a single adversarial example, rendering it similarly unsuitable for real-world applications.

For our adversarial effectiveness, we employ two metrics:  Fréchet Inception Distance (FID)~\cite{heusel2017gans} and Image Alignment (IA)~\cite{kumari2022multi} to measure the style similarity between the generated results and the reference images. Higher FID scores and lower IA values indicate greater dissimilarity, suggesting stronger adversarial effectiveness. It's important to note that while FID primarily assesses the diversity and quality of generated images, it may not fully capture the nuances of style generation. In contrast, IA is more suited for evaluating style generation tasks~\cite{kumari2022multi}.

As shown in Table~\ref{tab:fid}, our SITA method consistently achieves the best or second-best performance across all tasks for both FID and IA on the WikiArt~\cite{saleh2015large} and ArtBench~\cite{liao2022artbench} datasets, demonstrating its strong adversarial effectiveness and generalization.
Notably, in the T2I-Adapter task, SITA outperforms all other methods, achieving the best results for both FID and IA. Other methods that rely on the LDM architecture perform poorly, highlighting that our adversarial noise is not tied to specific LDM parameter versions but instead effectively disrupts the robust style features of the image.
In the Textual-Inversion and DreamBooth tasks, data poisoning methods like Anti-DreamBooth and Metacloak show better results than evasion attacks such as AdvDM and Glaze, due to their continuous parameter updates. However, these methods come at the cost of more noticeable adversarial noise and higher computational demands as shown in Table~\ref{tab:image_similarity}. In contrast, our SITA method strikes a balance between adversarial effectiveness, noise imperceptibility, and computational efficiency, achieving superior performance with minimal perceptible noise and lower computational costs.

\subsection{Transferability Analysis}

To assess the transferability of our method across different stylized generation models, we conduct a thorough evaluation of the adversarial transferability of two distinct style generation techniques: Textual-Inversion~\cite{gal2022image} and RB-Modulation~\cite{Rout2024RBModulationTP}. For Textual-Inversion, we employ several variants of Stable Diffusion (SD), including SD-v1.4, SD-v1.5, and SD-v2.1, to test the adversarial effectiveness of our approach across different versions of the model. For RB-Modulation, which uses the non-Stable Diffusion-based Würstchen~\cite{Pernias2023WuerstchenAE} model as its backbone, we examine the adversarial transferability of SITA to stylized generation methods based on non-Stable Diffusion models. Note that our method relies solely on a CLIP model to generate adversarial examples, for other methods, adversarial examples are generated using SD-v1.5 weights.

\begin{figure*}[t]
    \centering
    \includegraphics[width=\linewidth]{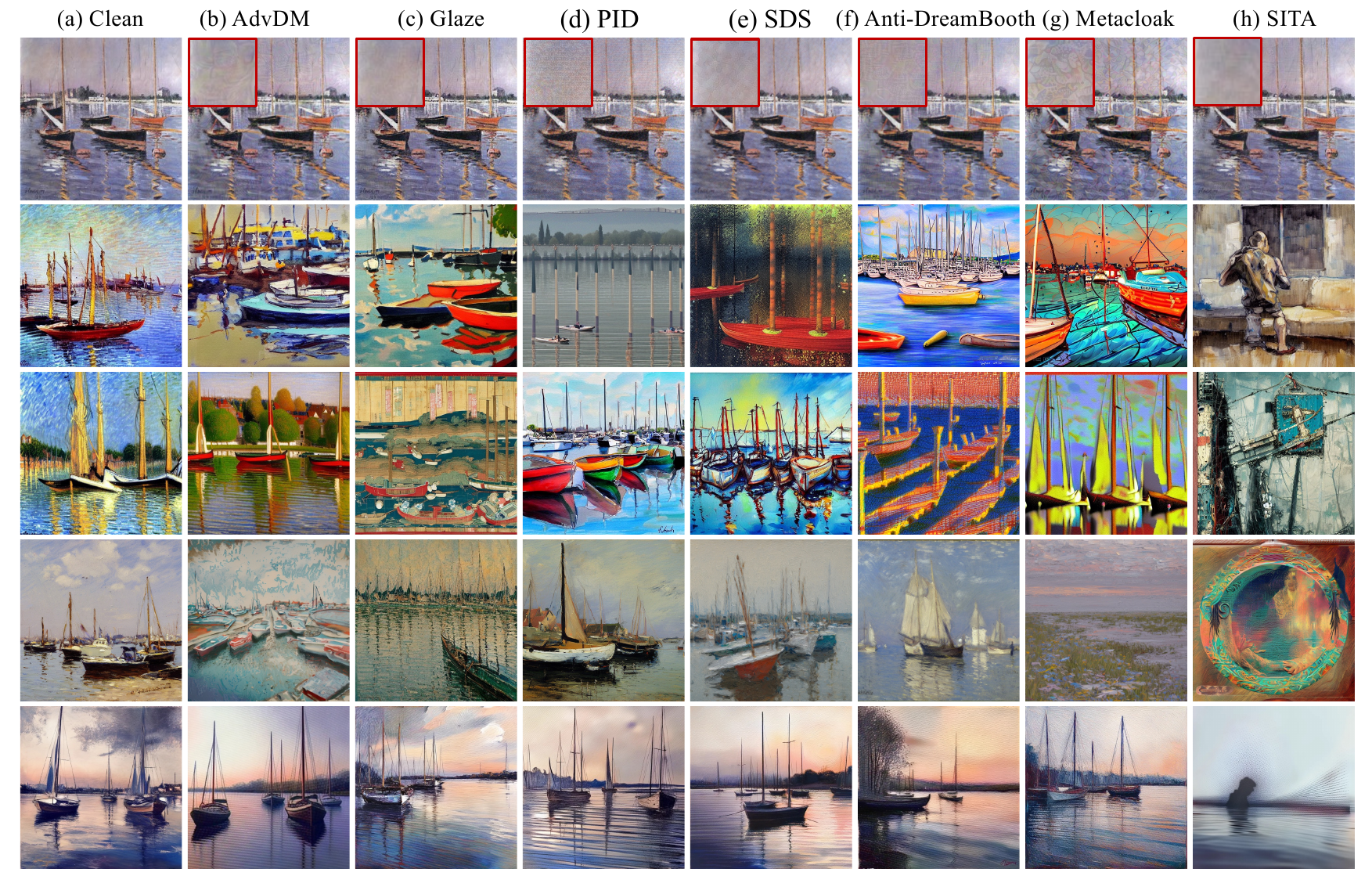}
    \caption{Transferability analysis results. The first row presents the reference images, while the second to fourth rows display the generation results from Textual-Inversion~\cite{gal2022image} using different versions of the Stable Diffusion model. The fifth row shows generation results from RB-Modulation~\cite{Rout2024RBModulationTP} based on the Würstchen~\cite{Pernias2023WuerstchenAE} model. Specifically, the second row corresponds to SD-v1.4, the third to SD-v1.5, the fourth to SD-v2.1, and the fifth to Würstchen.}
    \label{fig:trans}
\end{figure*}
\begin{table*}[t]
\centering
\setlength{\tabcolsep}{10pt}
\caption{Quantification results of FID \cite{heusel2017gans} and Image Alignment (IA) \cite{kumari2022multi} for transfer experiment on WikiArt~\cite{saleh2015large}.}
\begin{tabular}{l|cccccc|cc}
\toprule
\multirow{3}{*}{Method}                 & \multicolumn{6}{c|}{Textual-Inversion~\cite{gal2022image}}    & \multicolumn{2}{c}{RB-Modulation~\cite{Rout2024RBModulationTP}}   \\ \cmidrule(l){2-9} 
                                        & \multicolumn{2}{c}{SD-v1.4}       & \multicolumn{2}{c}{SD-v1.5}       & \multicolumn{2}{c|}{SD-v2.1}       & \multicolumn{2}{c}{Würstchen}\\ \cmidrule(l){2-9} 
                                        & FID$\uparrow$   & IA$\downarrow$ & FID$\uparrow$   & IA$\downarrow$ & FID$\uparrow$   & IA$\downarrow$    & FID$\uparrow$   & IA$\downarrow$\\ \midrule
AdvDM\cite{liang2023adversarial}        & 347.45          & 0.738          & 358.13          & 0.725          & {\ul 350.13}          & 0.761       &{\ul 305.27}          & 0.784\\
Glaze\cite{shan2023glaze}               & 334.09          & 0.746          & 347.02          & 0.691          & 348.68          & {\ul 0.695}       &260.22          & 0.814\\
PID\cite{li2024pid}                     & 309.13          & 0.754          & 338.61          & {\ul 0.683}          & 295.82          & 0.761             &299.84          & {\ul 0.712}\\
SDS\cite{xue2023toward}                 & 294.03          & 0.746          & 332.68          & 0.716          & 325.78          & 0.734             &298.47          & 0.722\\
Anti-Dreambooth\cite{van2023anti}       & \textbf{388.77} & {\ul 0.693}    & {\ul 359.09}    & 0.684    & 314.37          & 0.790             &278.95          & 0.819\\
Metacloak\cite{liu2024metacloak}        & 323.81          & 0.729          & 354.40          & 0.687          & 316.28          & 0.784             &273.78          & 0.822\\
Ours                                    & {\ul 351.36}    & \textbf{0.684} & \textbf{364.54} & \textbf{0.679} & \textbf{401.28} & \textbf{0.641}    &\textbf{351.18} & \textbf{0.627}\\ \bottomrule
\end{tabular}
\label{tab:trans}
\end{table*}

As shown in Fig.~\ref{fig:trans} and summarized in Table~\ref{tab:trans}, our approach generates adversarial examples with minimal perceptible noise while maintaining strong adversarial effectiveness across both Textual-Inversion and RB-Modulation, regardless of the model version. The generated images deviate significantly from the reference style, resulting in reduced overall quality. Our method consistently achieves the best or second-best FID scores across all versions, and the IA metric ranks highest in every case, demonstrating both the stability and transferability of our approach.
This superior performance can be attributed to two key factors: First, we use the CLIP image encoder, which offers a robust feature representation space that is independent of any specific generation model, allowing the adversarial examples to transfer effectively across different models. Second, the CLIP-based destylization loss effectively decouples the style information, hindering various stylized generation methods from capturing the underlying style generation patterns, thereby enhancing the transferability of the adversarial examples.

\begin{figure}[t]
    \centering
    \includegraphics[width=\linewidth]{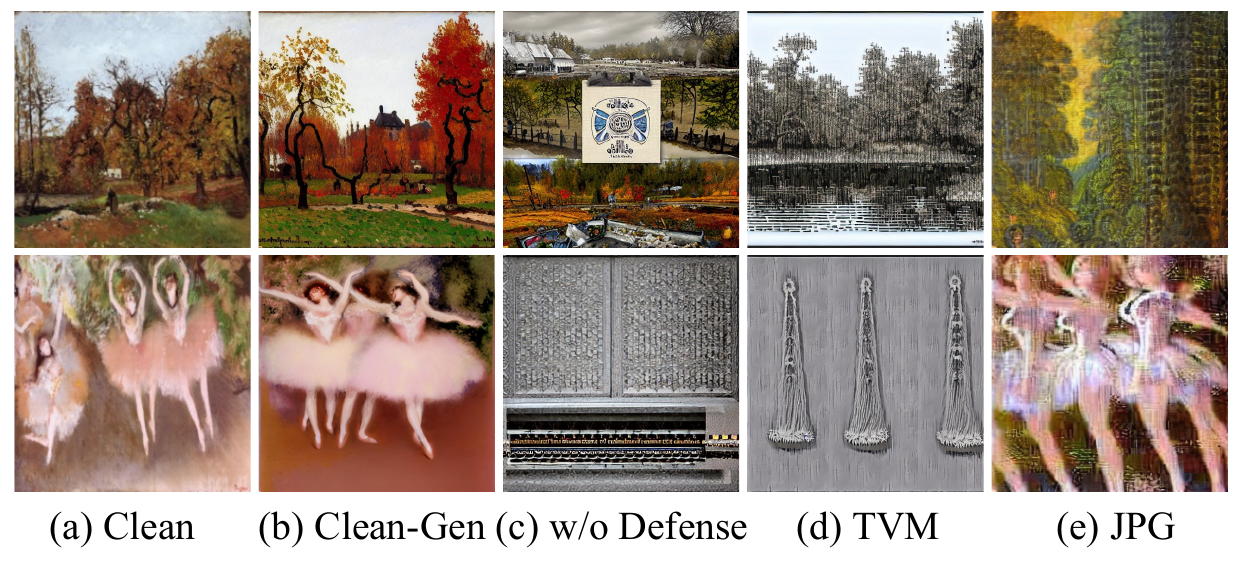}
    \caption{Visualization results under defense setting. (a) shows the clean sample, while (b) displays results generated using (a) as reference images. (c) presents the results with SITA's protection, and (d) and (e) show the outcomes after applying TVM defense\cite{Guo2018CounteringAI} and JPG compression defense\cite{Dziugaite2016ASO}.}
    \label{fig:defense}
\end{figure}
\begin{figure}[t]
    \centering
    \includegraphics[width=\linewidth]{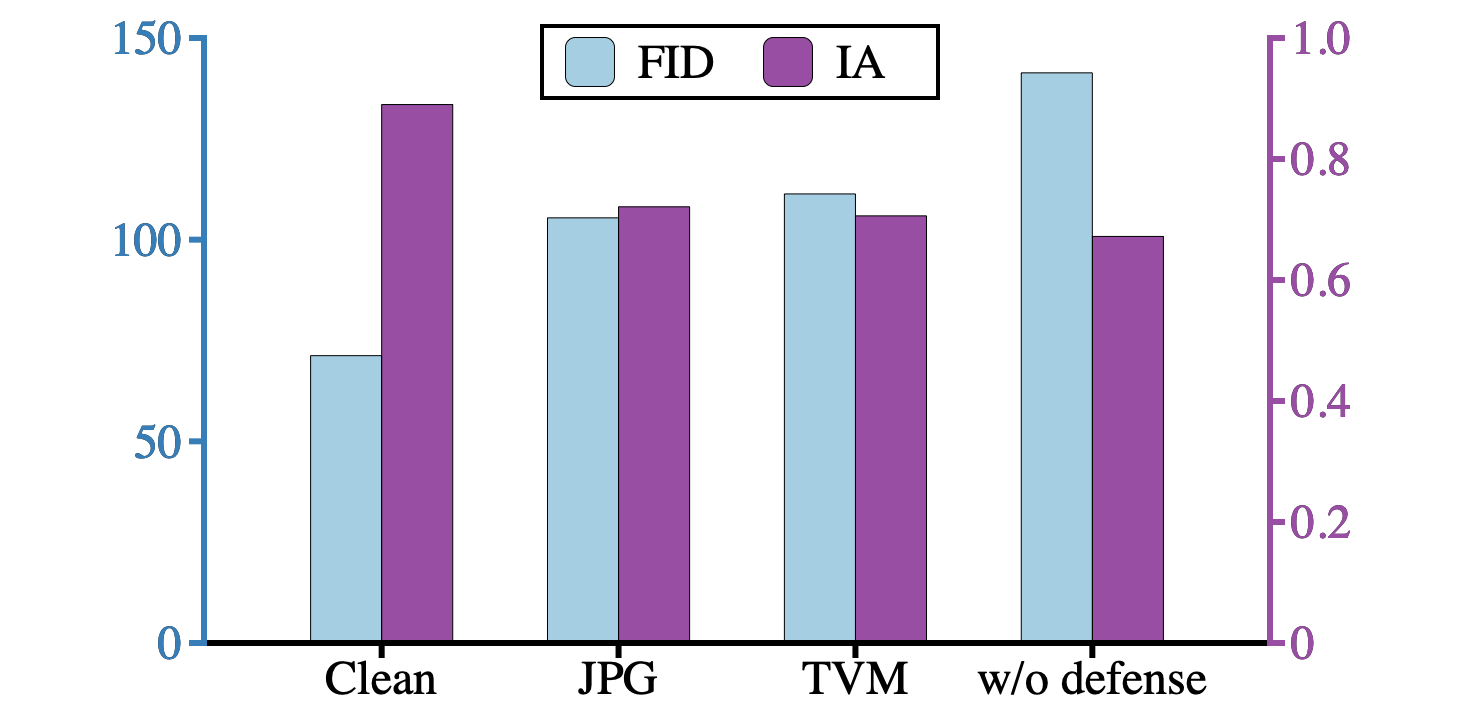}
    \caption{Quantification results for different defense settings on T2I-Adapter~\cite{mou2023t2i}.}
    \label{tab:defense}
\end{figure}
\subsection{Study on Robustness Against Defense Methods}
We also conduct both qualitative (See Fig.~\ref{fig:defense}) and quantitative (See Fig.~\ref{tab:defense}) experiments to assess the effectiveness of our SITA in scenarios with defensive measures based on T2I-Adapter~\cite{mou2023t2i}. Specifically, we evaluate its performance against two common preprocessing defenses: JPG compression defense \cite{Dziugaite2016ASO} and TVM defense \cite{Guo2018CounteringAI}. Note that the TVM defense includes image preprocessing methods such as Gaussian blur, Gaussian noise, and bit depth reduction, etc. 
From Fig.~\ref{fig:defense}, it can be observed that while the effectiveness of SITA is reduced under these defenses, the generated stylized images exhibit blurriness and meaningless artifacts, indicating that the model still fails to fully capture the reference image's style. Furthermore, as seen in Fig.~\ref{tab:defense}, the FID and IA metrics deteriorate compared to the case without defenses but remain better than those of clean samples.
This observation affirms that our proposed SITA retains a degree of robustness even when confronted with these specific defenses.

\subsection{Ablation Study}
\begin{figure}[t]
    \centering
    \includegraphics[width=\linewidth]{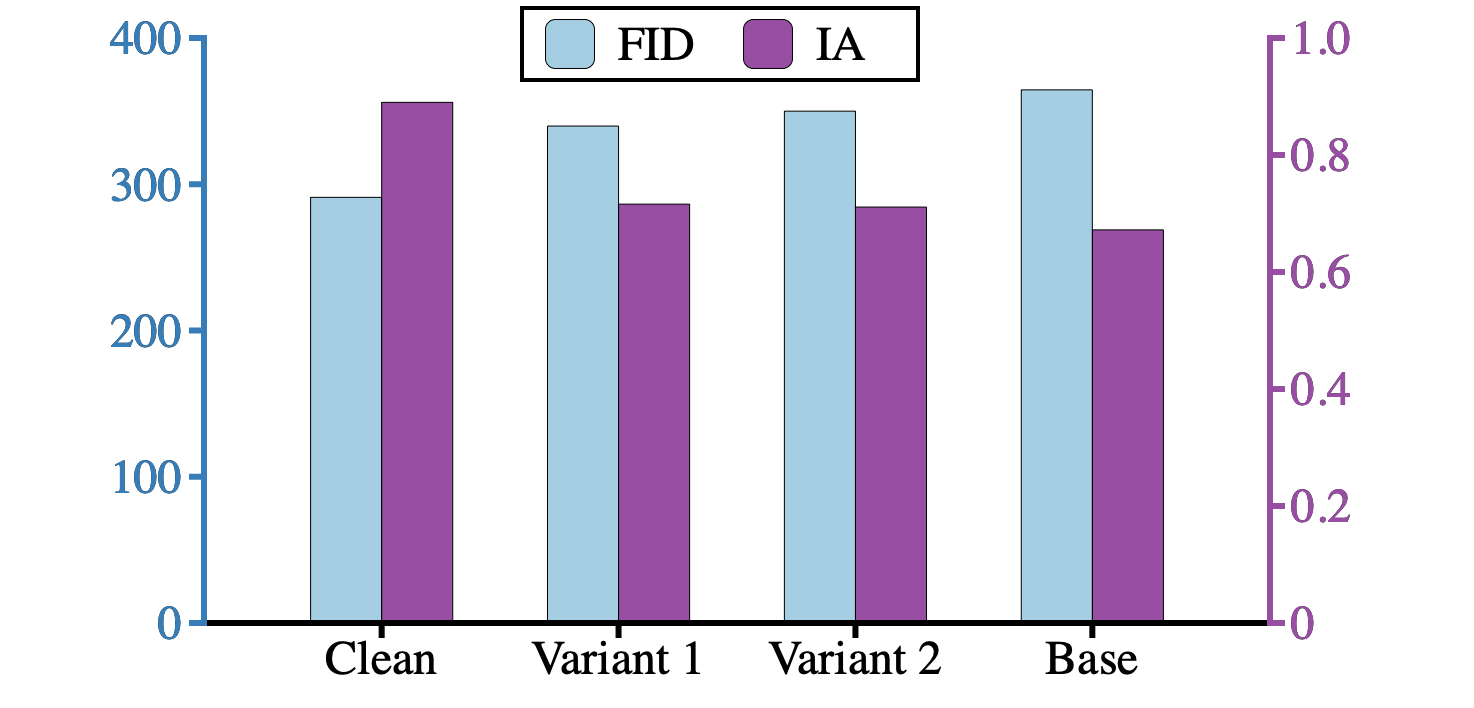}
    \caption{Quantification results for clip parameter settings on Textual-Inversion~\cite{gal2022image}.}
    \label{fig:clip}
\end{figure}

\noindent\textbf{Exploration of CLIP Variants.} We evaluate the impact of different CLIP image encoder versions on adversarial attack performance in the Textual-Inversion~\cite{gal2022image}. The quantitative results are shown in Fig.~\ref{fig:clip}, where variants 1 and 2 correspond to \textit{CLIP-ViT-Huge}\footnote{https://huggingface.co/laion/CLIP-ViT-H-14-laion2B-s32B-b79K} and \textit{CLIP-ViT-Base}\footnote{https://huggingface.co/openai/clip-vit-base-patch16}, respectively. The base model in our main experiment is \textit{CLIP-ViT-Large}\footnote{https://huggingface.co/openai/clip-vit-large-patch14}. The base variant shows the largest increase in FID and the greatest decrease in IA compared to clean samples, indicating the strongest adversarial effect.
Selecting other CLIP weight variants also effectively reduces style generation results using our SITA method, demonstrating that the CLIP-based destylization loss proposed by our approach is both effective and transferable across different CLIP image encoders. 

\begin{table}[t]
\caption{Ablation study on loss components.}
\label{tab:ablation}
\centering
\begin{tabularx}{0.48\textwidth}{@{} >{\centering\arraybackslash}p{0.35cm} >{\centering\arraybackslash}p{0.35cm} >{\centering\arraybackslash}p{0.5cm} >{\centering\arraybackslash}p{0.5cm} >{\centering\arraybackslash}p{0.5cm} |>{\centering\arraybackslash}p{0.47cm} >{\centering\arraybackslash}p{0.43cm} >{\centering\arraybackslash}p{0.5cm} >{\centering\arraybackslash}p{0.5cm} >{\centering\arraybackslash}p{0.5cm}@{}}
\toprule
$\mathrm{Gray}$ & $\mathrm{Blur}$  & $L_2$$\downarrow$  & SSIM$\uparrow$      & IA$\downarrow$          & $\mathcal{L}_{homo}$   & $\mathcal{L}_{stru}$   & $L_2$$\downarrow$  & SSIM$\uparrow$      & IA$\downarrow$             \\ \midrule
    &     & \textbf{1.002} & \textbf{0.998} & 0.858          &     &     & 7.428 & 0.924 & \textbf{0.671}          \\
\checkmark    &     & 4.801          & 0.961          & 0.761          & \checkmark    &     & 6.461 & 0.937 & {\ul 0.672}          \\
    & \checkmark    & 4.764          & 0.961          & {\ul 0.686}          &     & \checkmark    & {\ul 6.062} & {\ul 0.946} & 0.679          \\
\checkmark    & \checkmark    & {\ul 4.236}          & {\ul 0.972}          & \textbf{0.672} & \checkmark    & \checkmark    & \textbf{4.236} & \textbf{0.972} & 0.673 \\ \bottomrule
\end{tabularx}
\end{table}

\noindent\textbf{Ablation of Loss Function Components.} We conduct an ablation study on the $\mathrm{Blur}(\cdot)$ and $\mathrm{Gray}(\cdot)$ functions in the CLIP-based destylization loss, as well as the $\mathcal{L}_{homo}$ and $\mathcal{L}_{stru}$ components in the structure perception loss for the T2I-Adapter task. The results are summarized in Table~\ref{tab:ablation}. The experiment shows that $\mathrm{Blur}(\cdot)$ and $\mathrm{Gray}(\cdot)$ confirm that color and brushstrokes are indeed key components of style, and removing this information improves adversarial effectiveness, with the best performance achieved when both are combined. Similarly, the results demonstrate that $\mathcal{L}_{homo}$ effectively limits noise distribution in homogenous regions, while $\mathcal{L}_{stru}$ helps guide noise to the structural edges and details of the image. The combination of $\mathcal{L}_{homo}$ and $\mathcal{L}_{stru}$ makes the noise more imperceptible and enhances visual fidelity.

\noindent\textbf{Exploration of Noise Constraint Settings.} We conduct a comparative study to evaluate the perceptibility of noise across different constraint settings while maintaining the same level of attack effectiveness on the T2I-Adapter~\cite{mou2023t2i}. Specifically, we explore two noise constraints: an explicit noise budget, such as those used in PGD~\cite{goodfellow2014explaining}, and an implicit pixel constraint, where we replace our Structure Perception Loss with an L2 loss. For the explicit noise constraint, we set the perturbation budget $\epsilon = 0.015$, step size $\alpha = 0.002$, and 40 iterations. For the L2 constraint, we set the learning rate to 0.002 to ensure comparable attack effectiveness, keeping all other parameters constant.

The results, presented in Table~\ref{tab:cons_decouple}, reveal that the noise produced by the explicit budget constraint is more perceptible than that generated by the other two implicit noise control methods. This discrepancy arises because the explicit budget constraint focuses solely on optimizing the adversarial objective, without effectively guiding the noise toward concealment. Specifically, the explicit budget constraint relies on a threshold-based clipping operation to limit the noise amplitude, which compromises the fidelity of the adversarial examples. In contrast, the other two constraints balance both adversarial effectiveness and noise perceptibility during optimization, resulting in less perceptible noise. Notably, while the L2 constraint method improves image quality by minimizing the mean squared error in pixel space, it still produces more perceptible noise than our proposed dual-domain structure perception loss at the same attack effectiveness. This demonstrates that our carefully designed dual-domain structure perception loss more effectively guides noise optimization towards imperceptibility, aligning better with human visual perception.

\begin{table}[t]
\centering
\caption{Ablation study on different noise constraints (left) and destylization strategies (right).}
\label{tab:cons_decouple}
\begin{tabularx}{0.98\linewidth}{@{} 
>{\centering\arraybackslash}p{1.3cm} 
>{\centering\arraybackslash}p{0.5cm} 
>{\centering\arraybackslash}p{0.5cm} 
>{\centering\arraybackslash}p{0.5cm} 
| 
>{\centering\arraybackslash}p{1.3cm} 
>{\centering\arraybackslash}p{0.5cm} 
>{\centering\arraybackslash}p{0.5cm} 
>{\centering\arraybackslash}p{0.5cm} 
@{}}
\toprule
\textbf{Constraint} & $L_2$$\downarrow$ & SSIM$\uparrow$ & IA$\downarrow$ & \textbf{Strategy} & $L_2$$\downarrow$ & SSIM$\uparrow$ & IA$\downarrow$ \\
\midrule
Budget         & 5.615               & 0.942              & 0.675             & $\mathcal{L}_{destyle}^\mathrm{a}$    & 5.049               & 0.942              & 0.675 \\
L2                      & 4.931               & 0.958              & 0.678             & $\mathcal{L}_{destyle}^\mathrm{b}$    & 4.516               & 0.971              & 0.768 \\
SITA & \textbf{4.236}      & \textbf{0.972}     & \textbf{0.673}    & $\mathcal{L}_{destyle}^\mathrm{SITA}$ & \textbf{4.236}      & \textbf{0.972}     & \textbf{0.673} \\
\bottomrule
\end{tabularx}
\end{table}
\begin{figure*}[t]  
    \centering
    \begin{minipage}[b]{0.325\textwidth}
        \centering
        \includegraphics[width=\textwidth]{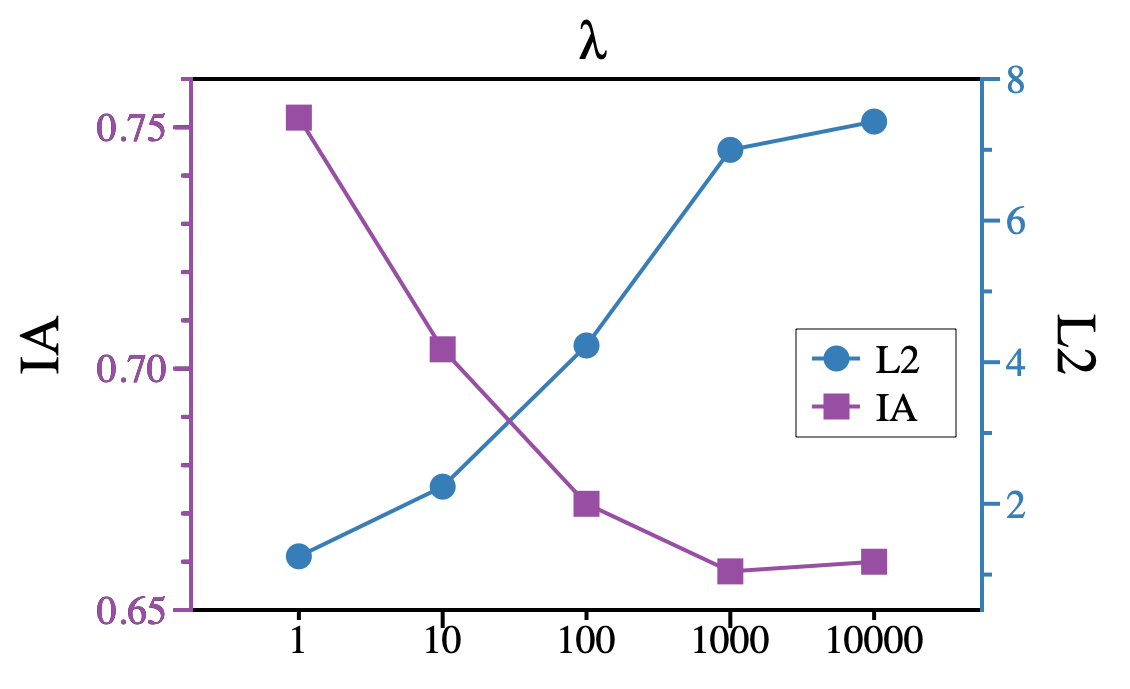}
        \parbox[t]{\textwidth}{\centering \small (a) Ablation Study on $\lambda$. } 
    \end{minipage}
    \hfill
    \begin{minipage}[b]{0.325\textwidth}
        \centering
        \includegraphics[width=\textwidth]{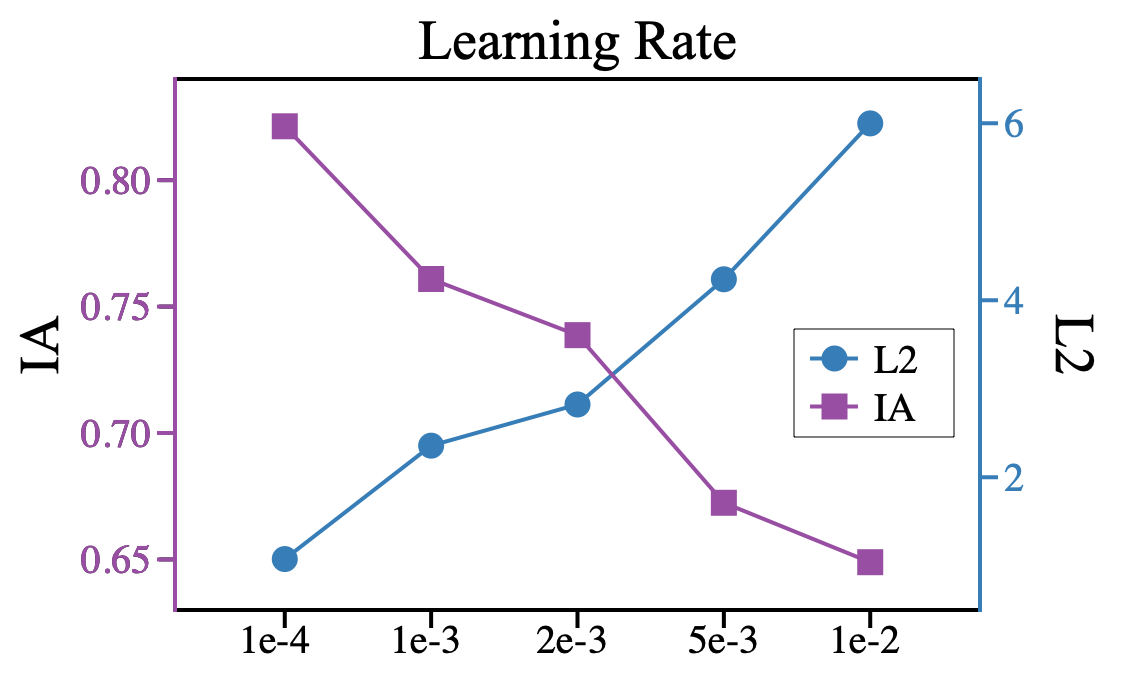}
        \parbox[t]{\textwidth}{\centering \small (b) Ablation Study on learning rate.} 
    \end{minipage}
    \hfill
    \begin{minipage}[b]{0.325\textwidth}
        \centering
        \includegraphics[width=\textwidth]{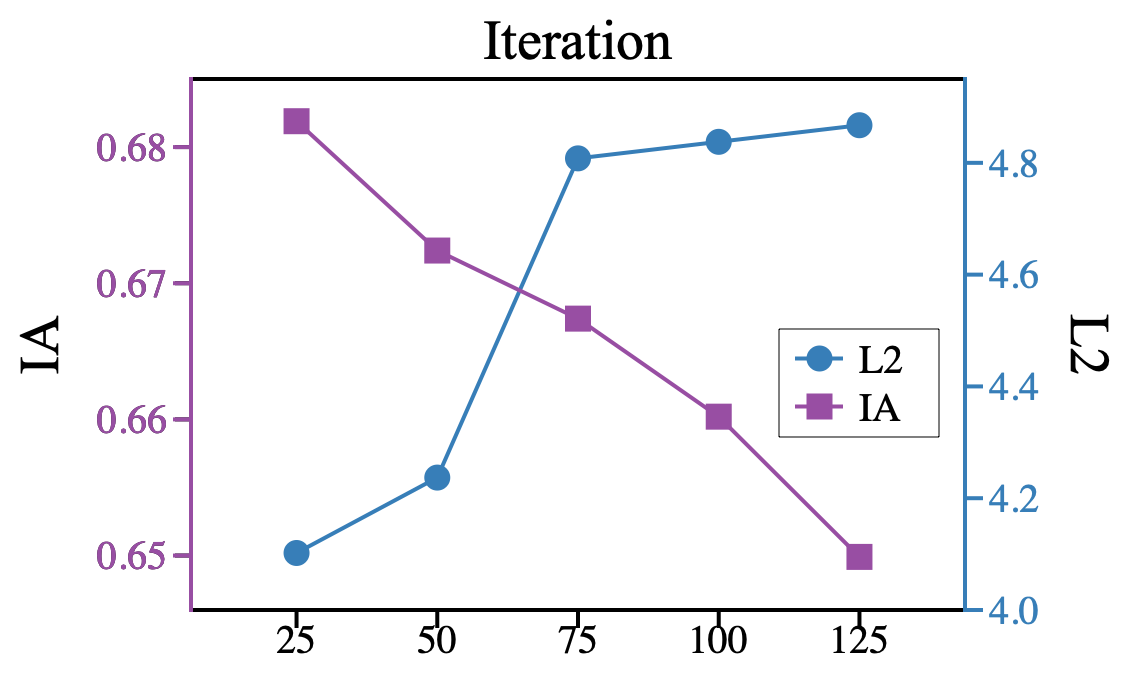}
        \parbox[t]{\textwidth}{\centering \small (c) Ablation Study on iteration.} 
    \end{minipage}
    \caption{Ablation study on the impact of $\lambda$, learning rate, and iteration settings in the T2I-Adapter\cite{mou2023t2i} task.}
    \label{fig:hyper}
\end{figure*}

\noindent\textbf{Exploration of destylization strategies.}
Beyond the Destylization Loss proposed in this paper, we also explore several alternative formulations, namely: $ \mathcal{L}_{destyle}^\mathrm{a} = -||E(X_s)-E(X_{adv})||_{1} $, which directly maximizes the $L_{1}$ distance between the CLIP embeddings of the original style reference image and the adversarial example;  $ \mathcal{L}_{destyle}^\mathrm{b} = -||\mathrm{Cos}(E(X_s), E(X_c))-\mathrm{Cos}(E(X_{adv}), E(X_c))||_{1} $, which first computes the style distance using cosine similarity and then maximizes the distance between the two style distances as the Destylization Loss.
The results, presented in Table~\ref{tab:cons_decouple}, indicate that while variant $\mathcal{L}_{destyle}^\mathrm{a}$can still disrupt the style generation process, it requires a significantly larger noise budget due to their lack of explicit style decoupling. In contrast, variant $\mathcal{L}_{destyle}^\mathrm{b}$ struggles to effectively degrade the stylization results. We hypothesize that this limitation arises because, in computing the style distance, cosine similarity is reduced to a scalar value, whereas our original destylization loss operates in a vector space. By compressing complex style representations into a single scalar, critical style information is lost, thereby weakening the ability to generate effective destylized adversarial examples.
These findings further validate the effectiveness of our proposed destylization loss. By carefully designing a targeted distance formulation, our method successfully extracts and disentangles style information, leading to more efficient and transferable adversarial examples.

\noindent\textbf{Exploration of Hyperparameters Settings.} We conduct our ablation studies on the T2I-Adapter \cite{mou2023t2i}, focusing initially on the impact of the coefficient term $\lambda$ in $L_{destyle}$. As detailed in Fig.~\ref{fig:hyper}(a), with a fixed learning rate of 0.005 and 50 iterations, varying $\lambda$ demonstrates that increasing values lead to increased perturbations and stronger adversarial effects. Thus, $\lambda$ is set to 100 to balance image quality and adversarial effectiveness optimally. Additionally, we investigate the impact of learning rate and iteration count on the perturbations generated by SITA, as shown in Fig.~\ref{fig:hyper}(b) and Fig.~\ref{fig:hyper}(c). First, we vary the learning rate while keeping the iteration count fixed at 50, and then we held the learning rate at 0.005 while changing the number of iterations, evaluating the effects in each case. Notably, increasing either parameter enhances both the visibility of noise and the adversarial effectiveness. Thus, we select a learning rate of 0.005 and 50 iterations as the optimal configuration, balancing noise perceptibility and adversarial strength.

\section{Conclusion}
In this paper, we introduce the Structurally Imperceptible and Transferable (SITA) adversarial attack, a novel approach tailored for stylized image generation. We explain the rationale behind adversarial attacks on image style information and present two key innovations: the CLIP-based destylization loss and the structure perception loss. These techniques allow the generation of adversarial examples with low computational cost, effectively preventing the extraction and replication of artwork styles while preserving high visual quality. Additionally, SITA exhibits robust and transferable adversarial effects across various models. Extensive experiments validate the efficacy of our approach.

Despite its strong performance across various stylized generation techniques and diffusion models, SITA does have certain limitations. Specifically, the CLIP-based Destylization Loss operates primarily through high-level style semantic disentanglement, which makes it less suitable for tasks that require precise texture replication or the preservation of localized patterns~\cite{chung2024style, hamazaspyan2023diffusion, zhang2022exact}. As a result, its applicability in these scenarios may be limited, requiring careful consideration. Addressing this limitation offers a valuable direction for future research, potentially by incorporating additional low-level feature constraints to mitigate the effects of texture imitation and further improve its versatility.

\small{
\bibliographystyle{IEEEtran}
\bibliography{refs}

\begin{thebibliography}{10}
\providecommand{\url}[1]{#1}
\csname url@samestyle\endcsname
\providecommand{\newblock}{\relax}
\providecommand{\bibinfo}[2]{#2}
\providecommand{\BIBentrySTDinterwordspacing}{\spaceskip=0pt\relax}
\providecommand{\BIBentryALTinterwordstretchfactor}{4}
\providecommand{\BIBentryALTinterwordspacing}{\spaceskip=\fontdimen2\font plus
\BIBentryALTinterwordstretchfactor\fontdimen3\font minus
  \fontdimen4\font\relax}
\providecommand{\BIBforeignlanguage}[2]{{%
\expandafter\ifx\csname l@#1\endcsname\relax
\typeout{** WARNING: IEEEtran.bst: No hyphenation pattern has been}%
\typeout{** loaded for the language `#1'. Using the pattern for}%
\typeout{** the default language instead.}%
\else
\language=\csname l@#1\endcsname
\fi
#2}}
\providecommand{\BIBdecl}{\relax}
\BIBdecl

\bibitem{sohl2015deep}
J.~Sohl-Dickstein, E.~Weiss, N.~Maheswaranathan, and S.~Ganguli, ``Deep
  unsupervised learning using nonequilibrium thermodynamics,'' in
  \emph{International Conference on Machine Learning}.\hskip 1em plus 0.5em
  minus 0.4em\relax PMLR, 2015, pp. 2256--2265.

\bibitem{ho2020denoising}
J.~Ho, A.~Jain, and P.~Abbeel, ``Denoising diffusion probabilistic models,''
  \emph{Advances in Neural Information Processing Systems}, vol.~33, pp.
  6840--6851, 2020.

\bibitem{ruiz2022dreambooth}
N.~Ruiz, Y.~Li, V.~Jampani, Y.~Pritch, M.~Rubinstein, and K.~Aberman,
  ``Dreambooth: Fine tuning text-to-image diffusion models for subject-driven
  generation,'' \emph{arXiv preprint arXiv:2208.12242}, 2022.

\bibitem{mou2023t2i}
C.~Mou, X.~Wang, L.~Xie, J.~Zhang, Z.~Qi, Y.~Shan, and X.~Qie, ``T2i-adapter:
  Learning adapters to dig out more controllable ability for text-to-image
  diffusion models,'' \emph{arXiv preprint arXiv:2302.08453}, 2023.

\bibitem{gal2022image}
R.~Gal, Y.~Alaluf, Y.~Atzmon, O.~Patashnik, A.~H. Bermano, G.~Chechik, and
  D.~Cohen-or, ``An image is worth one word: Personalizing text-to-image
  generation using textual inversion,'' in \emph{The Eleventh International
  Conference on Learning Representations}, 2022.

\bibitem{rombach2022high}
R.~Rombach, A.~Blattmann, D.~Lorenz, P.~Esser, and B.~Ommer, ``High-resolution
  image synthesis with latent diffusion models,'' in \emph{IEEE International
  Conference on Computer Vision}, 2022, pp. 10\,684--10\,695.

\bibitem{Szegedy2013IntriguingPO}
C.~Szegedy, W.~Zaremba, I.~Sutskever, J.~Bruna, D.~Erhan, I.~J. Goodfellow, and
  R.~Fergus, ``Intriguing properties of neural networks,'' \emph{CoRR}, vol.
  abs/1312.6199, 2013.

\bibitem{Goldblum2020DatasetSF}
M.~Goldblum, D.~Tsipras, C.~Xie, X.~Chen, A.~Schwarzschild, D.~X. Song,
  A.~Madry, B.~Li, and T.~Goldstein, ``Dataset security for machine learning:
  Data poisoning, backdoor attacks, and defenses,'' \emph{IEEE Transactions on
  Pattern Analysis and Machine Intelligence}, vol.~45, pp. 1563--1580, 2020.

\bibitem{liang2023adversarial}
C.~Liang, X.~Wu, Y.~Hua, J.~Zhang, Y.~Xue, T.~Song, Z.~Xue, R.~Ma, and H.~Guan,
  ``Adversarial example does good: Preventing painting imitation from diffusion
  models via adversarial examples,'' in \emph{International Conference on
  Machine Learning}, 2023, pp. 20\,763--20\,786.

\bibitem{xue2023toward}
H.~Xue, C.~Liang, X.~Wu, and Y.~Chen, ``Toward effective protection against
  diffusion-based mimicry through score distillation,'' in \emph{The Twelfth
  International Conference on Learning Representations}, 2023.

\bibitem{shan2023glaze}
S.~Shan, J.~Cryan, E.~Wenger, H.~Zheng, R.~Hanocka, and B.~Y. Zhao, ``Glaze:
  Protecting artists from style mimicry by {Text-to-Image} models,'' in
  \emph{32nd USENIX Security Symposium}, 2023, pp. 2187--2204.

\bibitem{li2024pid}
A.~Li, Y.~Mo, M.~Li, and Y.~Wang, ``Pid: prompt-independent data protection
  against latent diffusion models,'' \emph{arXiv preprint arXiv:2406.15305},
  2024.

\bibitem{van2023anti}
T.~Van~Le, H.~Phung, T.~H. Nguyen, Q.~Dao, N.~N. Tran, and A.~Tran,
  ``Anti-dreambooth: Protecting users from personalized text-to-image
  synthesis,'' in \emph{IEEE International Conference on Computer Vision},
  2023, pp. 2116--2127.

\bibitem{liu2024metacloak}
Y.~Liu, C.~Fan, Y.~Dai, X.~Chen, P.~Zhou, and L.~Sun, ``Metacloak: Preventing
  unauthorized subject-driven text-to-image diffusion-based synthesis via
  meta-learning,'' in \emph{Proceedings of the IEEE/CVF Conference on Computer
  Vision and Pattern Recognition}, 2024, pp. 24\,219--24\,228.

\bibitem{radford2021learning}
A.~Radford, J.~W. Kim, C.~Hallacy, A.~Ramesh, G.~Goh, S.~Agarwal, G.~Sastry,
  A.~Askell, P.~Mishkin, J.~Clark \emph{et~al.}, ``Learning transferable visual
  models from natural language supervision,'' in \emph{International conference
  on machine learning}, 2021, pp. 8748--8763.

\bibitem{xu2023stylerdalle}
Z.~Xu, E.~Sangineto, and N.~Sebe, ``Stylerdalle: Language-guided style transfer
  using a vector-quantized tokenizer of a large-scale generative model,'' in
  \emph{Proceedings of the IEEE/CVF International Conference on Computer
  Vision}, 2023, pp. 7601--7611.

\bibitem{chefer2022image}
H.~Chefer, S.~Benaim, R.~Paiss, and L.~Wolf, ``Image-based clip-guided essence
  transfer,'' in \emph{European Conference on Computer Vision}.\hskip 1em plus
  0.5em minus 0.4em\relax Springer, 2022, pp. 695--711.

\bibitem{ramesh2022hierarchical}
A.~Ramesh, P.~Dhariwal, A.~Nichol, C.~Chu, and M.~Chen, ``Hierarchical
  text-conditional image generation with clip latents,'' \emph{arXiv preprint
  arXiv:2204.06125}, vol.~1, no.~2, p.~3, 2022.

\bibitem{frans2022clipdraw}
K.~Frans, L.~Soros, and O.~Witkowski, ``Clipdraw: Exploring text-to-drawing
  synthesis through language-image encoders,'' \emph{Advances in Neural
  Information Processing Systems}, vol.~35, pp. 5207--5218, 2022.

\bibitem{song2020denoising}
J.~Song, C.~Meng, and S.~Ermon, ``Denoising diffusion implicit models,''
  \emph{arXiv preprint arXiv:2010.02502}, 2020.

\bibitem{deng2022stytr2}
Y.~Deng, F.~Tang, W.~Dong, C.~Ma, X.~Pan, L.~Wang, and C.~Xu, ``Stytr2: Image
  style transfer with transformers,'' in \emph{IEEE Conference on Computer
  Vision and Pattern Recognition}, 2022, pp. 11\,326--11\,336.

\bibitem{gatys2016image}
L.~A. Gatys, A.~S. Ecker, and M.~Bethge, ``Image style transfer using
  convolutional neural networks,'' in \emph{IEEE Conference on Computer Vision
  and Pattern Recognition}, 2016, pp. 2414--2423.

\bibitem{zhang2022domain}
Y.~Zhang, F.~Tang, W.~Dong, H.~Huang, C.~Ma, T.-Y. Lee, and C.~Xu, ``Domain
  enhanced arbitrary image style transfer via contrastive learning,'' in
  \emph{ACM SIGGRAPH Conference Proceedings}, 2022, pp. 1--8.

\bibitem{xiao2022appearance}
W.~Xiao, C.~Xu, J.~Mai, X.~Xu, Y.~Li, C.~Li, X.~Liu, and S.~He,
  ``Appearance-preserved portrait-to-anime translation via proxy-guided domain
  adaptation,'' \emph{IEEE Transactions on Visualization and Computer
  Graphics}, 2022.

\bibitem{ruiz2023dreambooth}
N.~Ruiz, Y.~Li, V.~Jampani, Y.~Pritch, M.~Rubinstein, and K.~Aberman,
  ``Dreambooth: Fine tuning text-to-image diffusion models for subject-driven
  generation,'' in \emph{IEEE Conference on Computer Vision and Pattern
  Recognition}, 2023, pp. 22\,500--22\,510.

\bibitem{goodfellow2014explaining}
I.~J. Goodfellow, J.~Shlens, and C.~Szegedy, ``Explaining and harnessing
  adversarial examples,'' \emph{arXiv preprint arXiv:1412.6572}, 2014.

\bibitem{madry2017towards}
A.~Madry, A.~Makelov, L.~Schmidt, D.~Tsipras, and A.~Vladu, ``Towards deep
  learning models resistant to adversarial attacks,'' \emph{arXiv preprint
  arXiv:1706.06083}, 2017.

\bibitem{Carlini2016TowardsET}
N.~Carlini and D.~A. Wagner, ``Towards evaluating the robustness of neural
  networks,'' \emph{2017 IEEE Symposium on Security and Privacy (SP)}, pp.
  39--57, 2016.

\bibitem{moosavi2016deepfool}
S.-M. Moosavi-Dezfooli, A.~Fawzi, and P.~Frossard, ``Deepfool: a simple and
  accurate method to fool deep neural networks,'' in \emph{IEEE conference on
  computer vision and pattern recognition}, 2016, pp. 2574--2582.

\bibitem{papernot2016limitations}
N.~Papernot, P.~McDaniel, S.~Jha, M.~Fredrikson, Z.~B. Celik, and A.~Swami,
  ``The limitations of deep learning in adversarial settings,'' in \emph{IEEE
  European symposium on security and privacy (EuroS\&P)}, 2016, pp. 372--387.

\bibitem{ma2023transferable}
H.~Ma, K.~Xu, X.~Jiang, Z.~Zhao, and T.~Sun, ``Transferable black-box attack
  against face recognition with spatial mutable adversarial patch,'' \emph{IEEE
  Transactions on Information Forensics and Security}, 2023.

\bibitem{weng2023logit}
J.~Weng, Z.~Luo, S.~Li, N.~Sebe, and Z.~Zhong, ``Logit margin matters:
  Improving transferable targeted adversarial attack by logit calibration,''
  \emph{IEEE Transactions on Information Forensics and Security}, vol.~18, pp.
  3561--3574, 2023.

\bibitem{chen2024diffusion}
J.~Chen, H.~Chen, K.~Chen, Y.~Zhang, Z.~Zou, and Z.~Shi, ``Diffusion models for
  imperceptible and transferable adversarial attack,'' \emph{IEEE Transactions
  on Pattern Analysis and Machine Intelligence}, 2024.

\bibitem{zhang2024perception}
S.~Zhang, B.~Zheng, P.~Jiang, L.~Zhao, C.~Shen, and Q.~Wang,
  ``Perception-driven imperceptible adversarial attack against decision-based
  black-box models,'' \emph{IEEE Transactions on Information Forensics and
  Security}, 2024.

\bibitem{luo2022frequency}
C.~Luo, Q.~Lin, W.~Xie, B.~Wu, J.~Xie, and L.~Shen, ``Frequency-driven
  imperceptible adversarial attack on semantic similarity,'' in \emph{IEEE
  Conference on Computer Vision and Pattern Recognition}, 2022, pp.
  15\,315--15\,324.

\bibitem{duan2021advdrop}
R.~Duan, Y.~Chen, D.~Niu, Y.~Yang, A.~K. Qin, and Y.~He, ``Advdrop: Adversarial
  attack to dnns by dropping information,'' in \emph{IEEE International
  Conference on Computer Vision}, 2021, pp. 7506--7515.

\bibitem{long2022frequency}
Y.~Long, Q.~Zhang, B.~Zeng, L.~Gao, X.~Liu, J.~Zhang, and J.~Song, ``Frequency
  domain model augmentation for adversarial attack,'' in \emph{European
  Conference on Computer Vision}, 2022, pp. 549--566.

\bibitem{zhou2023downstreamagnostic}
Z.~Zhou, S.~Hu, R.~Zhao, Q.~Wang, L.~Y. Zhang, J.~Hou, and H.~Jin,
  ``Downstream-agnostic adversarial examples,'' \emph{arXiv preprint
  arXiv:2307.12280}, 2023.

\bibitem{MoosaviDezfooli2016UniversalAP}
S.-M. Moosavi-Dezfooli, A.~Fawzi, O.~Fawzi, and P.~Frossard, ``Universal
  adversarial perturbations,'' \emph{IEEE Conference on Computer Vision and
  Pattern Recognition}, pp. 86--94, 2016.

\bibitem{patashnik2021styleclip}
O.~Patashnik, Z.~Wu, E.~Shechtman, D.~Cohen-Or, and D.~Lischinski, ``Styleclip:
  Text-driven manipulation of stylegan imagery,'' in \emph{IEEE International
  Conference on Computer Vision}, 2021, pp. 2085--2094.

\bibitem{kwon2022clipstyler}
G.~Kwon and J.~C. Ye, ``Clipstyler: Image style transfer with a single text
  condition,'' in \emph{Computer Vision and Pattern Recognition}, 2022, pp.
  18\,062--18\,071.

\bibitem{vinker2022clipasso}
Y.~Vinker, E.~Pajouheshgar, J.~Y. Bo, R.~C. Bachmann, A.~H. Bermano,
  D.~Cohen-Or, A.~Zamir, and A.~Shamir, ``Clipasso: Semantically-aware object
  sketching,'' \emph{ACM Transactions on Graphics}, vol.~41, no.~4, pp. 1--11,
  2022.

\bibitem{Gatys2015ANA}
L.~A. Gatys, A.~S. Ecker, and M.~Bethge, ``A neural algorithm of artistic
  style,'' \emph{ArXiv}, vol. abs/1508.06576, 2015.

\bibitem{Johnson2016PerceptualLF}
J.~Johnson, A.~Alahi, and L.~Fei-Fei, ``Perceptual losses for real-time style
  transfer and super-resolution,'' \emph{ArXiv}, vol. abs/1603.08155, 2016.

\bibitem{Huang2017ArbitraryST}
X.~Huang and S.~J. Belongie, ``Arbitrary style transfer in real-time with
  adaptive instance normalization,'' \emph{IEEE International Conference on
  Computer Vision}, pp. 1510--1519, 2017.

\bibitem{Wang2023StyleDiffusionCD}
Z.~Wang, L.~Zhao, and W.~Xing, ``Stylediffusion: Controllable disentangled
  style transfer via diffusion models,'' \emph{ArXiv}, vol. abs/2308.07863,
  2023.

\bibitem{Meyer1979-MEYTAT-4}
L.~B. Meyer, ``Toward a theory of style,'' in \emph{The Concept of Style},
  L.~B. Meyer and B.~Lang, Eds., 1979, pp. 3--44.

\bibitem{Deac2006FeatureSF}
A.~I. Deac, J.~C.~A. van~der Lubbe, and E.~Backer, ``Feature selection for
  paintings classification by optimal tree pruning,'' in \emph{Multimedia
  Content Representation, Classification and Security}, 2006.

\bibitem{Zujovic2009ClassifyingPB}
J.~Zujovic, L.~M. Gandy, S.~E. Friedman, B.~Pardo, and T.~N. Pappas,
  ``Classifying paintings by artistic genre: An analysis of features \&
  classifiers,'' \emph{IEEE International Workshop on Multimedia Signal
  Processing}, pp. 1--5, 2009.

\bibitem{Agarwal2015GenreAS}
S.~Agarwal, H.~Karnick, N.~Pant, and U.~Patel, ``Genre and style based painting
  classification,'' \emph{IEEE Winter Conference on Applications of Computer
  Vision}, pp. 588--594, 2015.

\bibitem{shensa1992discrete}
M.~J. Shensa \emph{et~al.}, ``The discrete wavelet transform: wedding the a
  trous and mallat algorithms,'' \emph{IEEE Transactions on signal processing},
  vol.~40, no.~10, pp. 2464--2482, 1992.

\bibitem{saleh2015large}
B.~Saleh and A.~Elgammal, ``Large-scale classification of fine-art paintings:
  Learning the right metric on the right feature,'' \emph{arXiv preprint
  arXiv:1505.00855}, 2015.

\bibitem{liao2022artbench}
P.~Liao, X.~Li, X.~Liu, and K.~Keutzer, ``The artbench dataset: Benchmarking
  generative models with artworks,'' \emph{arXiv preprint arXiv:2206.11404},
  2022.

\bibitem{paszke2019pytorch}
A.~Paszke, S.~Gross, F.~Massa, A.~Lerer, J.~Bradbury, G.~Chanan, T.~Killeen,
  Z.~Lin, N.~Gimelshein, L.~Antiga \emph{et~al.}, ``Pytorch: An imperative
  style, high-performance deep learning library,'' \emph{Advances in neural
  information processing systems}, vol.~32, 2019.

\bibitem{heusel2017gans}
M.~Heusel, H.~Ramsauer, T.~Unterthiner, B.~Nessler, and S.~Hochreiter, ``Gans
  trained by a two time-scale update rule converge to a local nash
  equilibrium,'' \emph{Advances in neural information processing systems},
  vol.~30, 2017.

\bibitem{kumari2022multi}
N.~Kumari, B.~Zhang, R.~Zhang, E.~Shechtman, and J.-Y. Zhu, ``Multi-concept
  customization of text-to-image diffusion,'' \emph{arXiv preprint
  arXiv:2212.04488}, 2022.

\bibitem{wang2004image}
Z.~Wang, A.~C. Bovik, H.~R. Sheikh, and E.~P. Simoncelli, ``Image quality
  assessment: from error visibility to structural similarity,'' \emph{IEEE
  transactions on image processing}, vol.~13, no.~4, pp. 600--612, 2004.

\bibitem{gonzalez2008digital}
R.~C. Gonzalez and R.~E. Woods, \emph{Digital Image Processing}.\hskip 1em plus
  0.5em minus 0.4em\relax Prentice Hall, 2008.

\bibitem{Rout2024RBModulationTP}
L.~Rout, Y.~Chen, N.~Ruiz, A.~Kumar, C.~Caramanis, S.~Shakkottai, and W.-S.
  Chu, ``Rb-modulation: Training-free personalization of diffusion models using
  stochastic optimal control,'' \emph{ArXiv}, vol. abs/2405.17401, 2024.

\bibitem{Pernias2023WuerstchenAE}
P.~Pernias, D.~Rampas, M.~L. Richter, C.~Pal, and M.~Aubreville, ``Wuerstchen:
  An efficient architecture for large-scale text-to-image diffusion models,''
  2023.

\bibitem{Guo2018CounteringAI}
C.~Guo, M.~Rana, M.~Ciss{\'e}, and L.~van~der Maaten, ``Countering adversarial
  images using input transformations,'' \emph{ArXiv}, vol. abs/1711.00117,
  2018.

\bibitem{Dziugaite2016ASO}
G.~K. Dziugaite, Z.~Ghahramani, and D.~M. Roy, ``A study of the effect of jpg
  compression on adversarial images,'' \emph{ArXiv}, vol. abs/1608.00853, 2016.

\bibitem{chung2024style}
J.~Chung, S.~Hyun, and J.-P. Heo, ``Style injection in diffusion: A
  training-free approach for adapting large-scale diffusion models for style
  transfer,'' in \emph{Proceedings of the IEEE/CVF conference on computer
  vision and pattern recognition}, 2024, pp. 8795--8805.

\bibitem{hamazaspyan2023diffusion}
M.~Hamazaspyan and S.~Navasardyan, ``Diffusion-enhanced patchmatch: A framework
  for arbitrary style transfer with diffusion models,'' in \emph{Proceedings of
  the IEEE/CVF Conference on Computer Vision and Pattern Recognition}, 2023,
  pp. 797--805.

\bibitem{zhang2022exact}
Y.~Zhang, M.~Li, R.~Li, K.~Jia, and L.~Zhang, ``Exact feature distribution
  matching for arbitrary style transfer and domain generalization,'' in
  \emph{Proceedings of the IEEE/CVF conference on computer vision and pattern
  recognition}, 2022, pp. 8035--8045.

\end{thebibliography}
}

\begin{IEEEbiography}[{\includegraphics[width=1in,height=1.25in,clip,keepaspectratio]{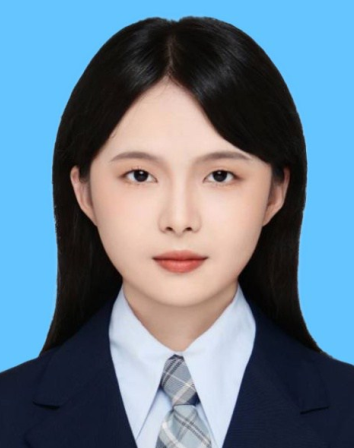}}]{Jingdan Kang}
is a graduate student at the School of Future Technology, South China University of Technology. She received her B.Sc. degree from South China University of Technology in 2023. Her research interests include privacy and security in computer vision.
\end{IEEEbiography}
\begin{IEEEbiography}[{\includegraphics[width=1in,height=1.25in,clip,keepaspectratio]{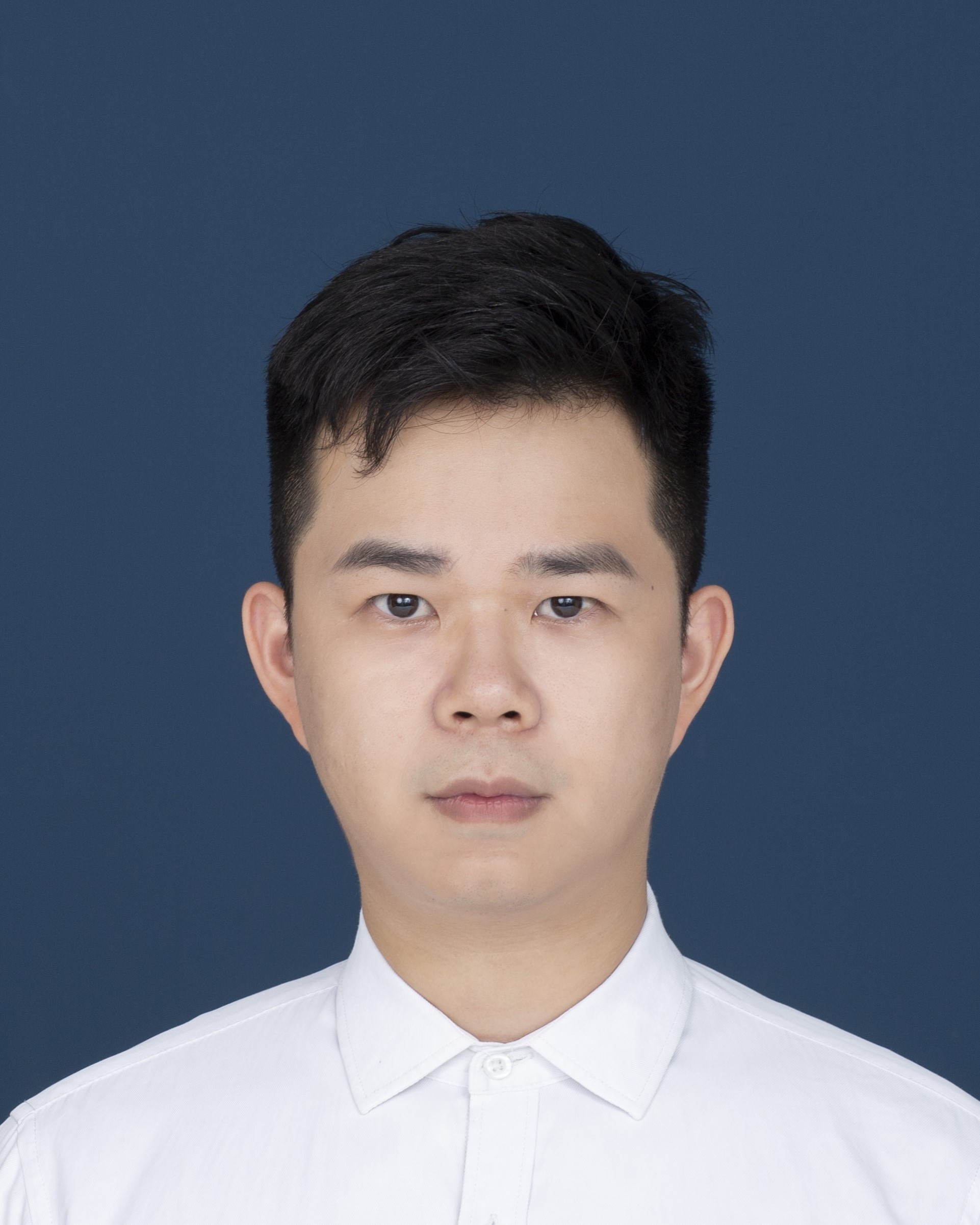}}]{Haoxin Yang} is a Ph.D. student at School of Computer Science \& Engineering, South China University of Technology. He obtained his B.Sc. and M.Sc. degrees from South China Agricultural University and Shenzhen University in 2019 and 2022, respectively. His research interests include privacy and security in computer vision and AIGC.\end{IEEEbiography}
\begin{IEEEbiography}[{\includegraphics[width=1in,height=1.25in,clip,keepaspectratio]{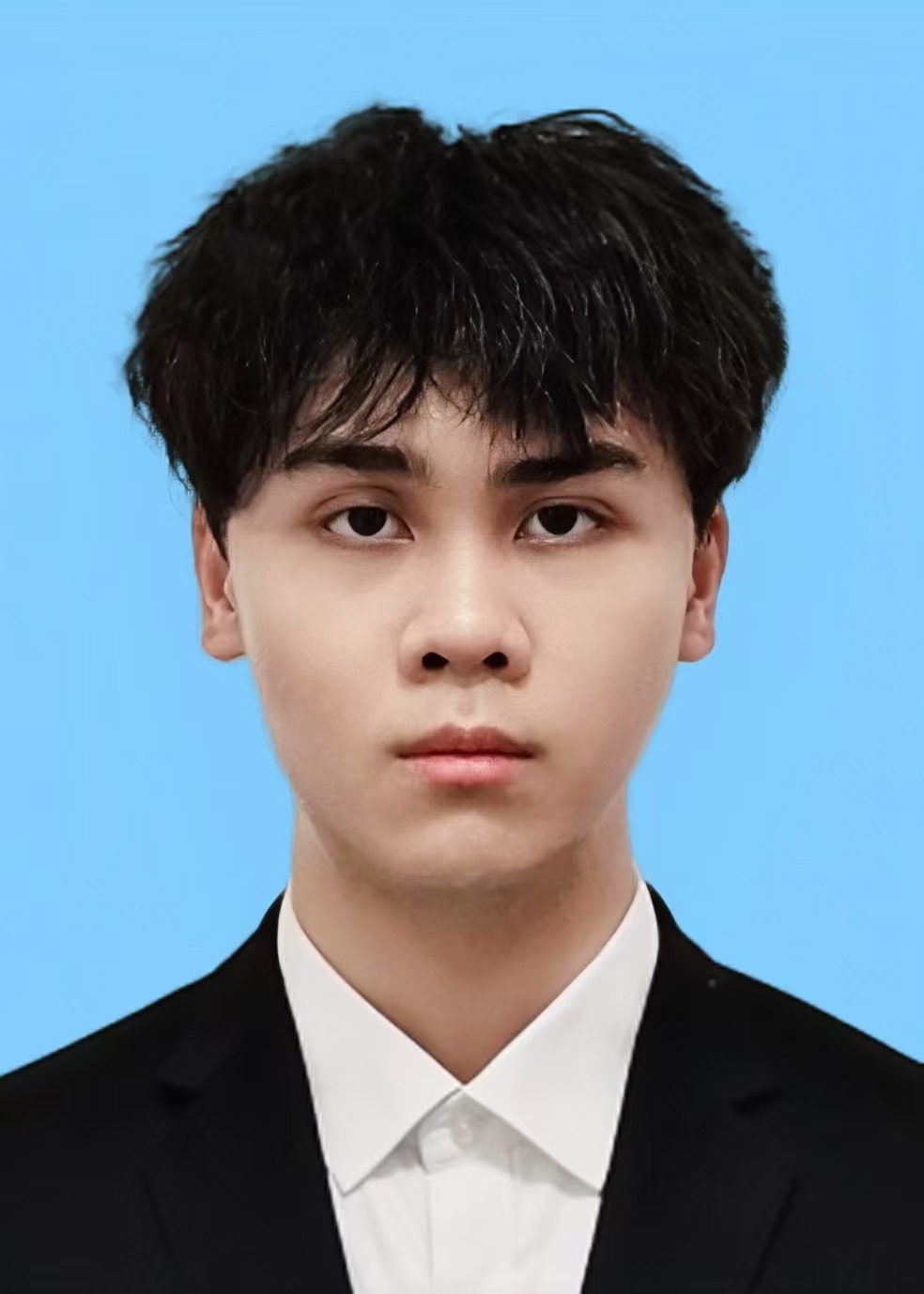}}]{Yan Cai}  is an undergraduate student at the School of Future Technology, South China University of Technology, enrolled in 2022. His research interests include privacy and security in computer vision.\end{IEEEbiography}
\begin{IEEEbiography}[{\includegraphics[width=1in,height=1.25in,clip,keepaspectratio]{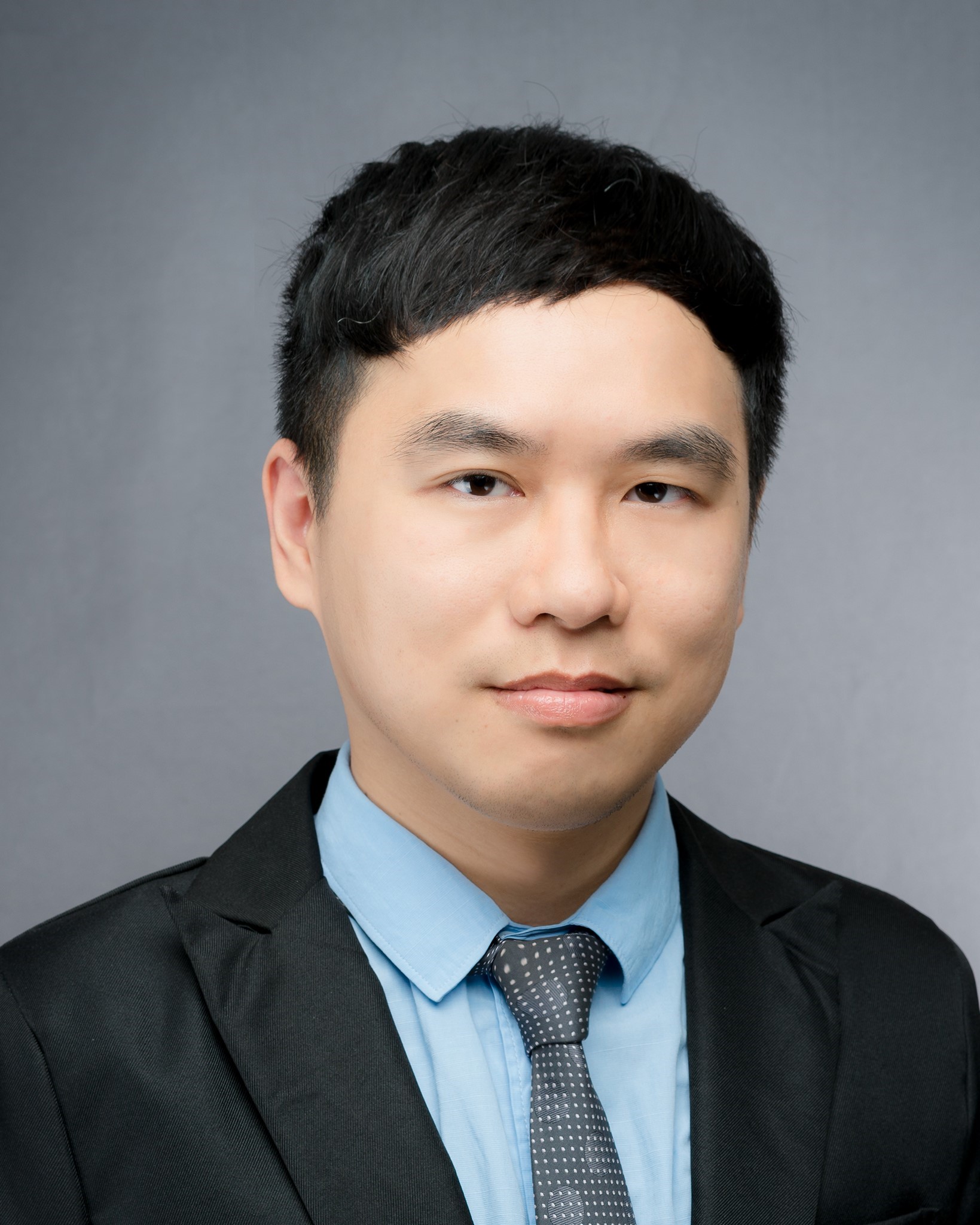}}]{Huaidong Zhang} 
is an Associate Professor in the School of Future Technology, South China University of Technology. He was a Postdoctoral Fellow at The Hong Kong Polytechnic University. He received his B.Eng. and Ph.D. degrees in Computer Science and Engineering from the South China University of Technology in 2015 and 2020 respectively. His research interests include computer vision, image processing, computer graphics and deep learning.\end{IEEEbiography}
\begin{IEEEbiography}[{\includegraphics[width=1in,height=1.25in,clip,keepaspectratio]{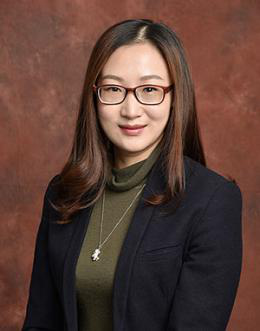}}]{Xuemiao Xu} received her B.S. and M.S. degrees in Computer Science and Engineering from South China University of Technology in 2002 and 2005 respectively, and Ph.D. degree in Computer Science and Engineering from The Chinese University of Hong Kong in 2009. She is currently a professor in the School of Computer Science and Engineering, South China University of Technology. Her research interests include object detection, tracking, recognition, and image, video understanding and synthesis, particularly their applications in the intelligent transportation.
\end{IEEEbiography}
\begin{IEEEbiography}[{\includegraphics[width=1in,height=1.25in,clip,keepaspectratio]{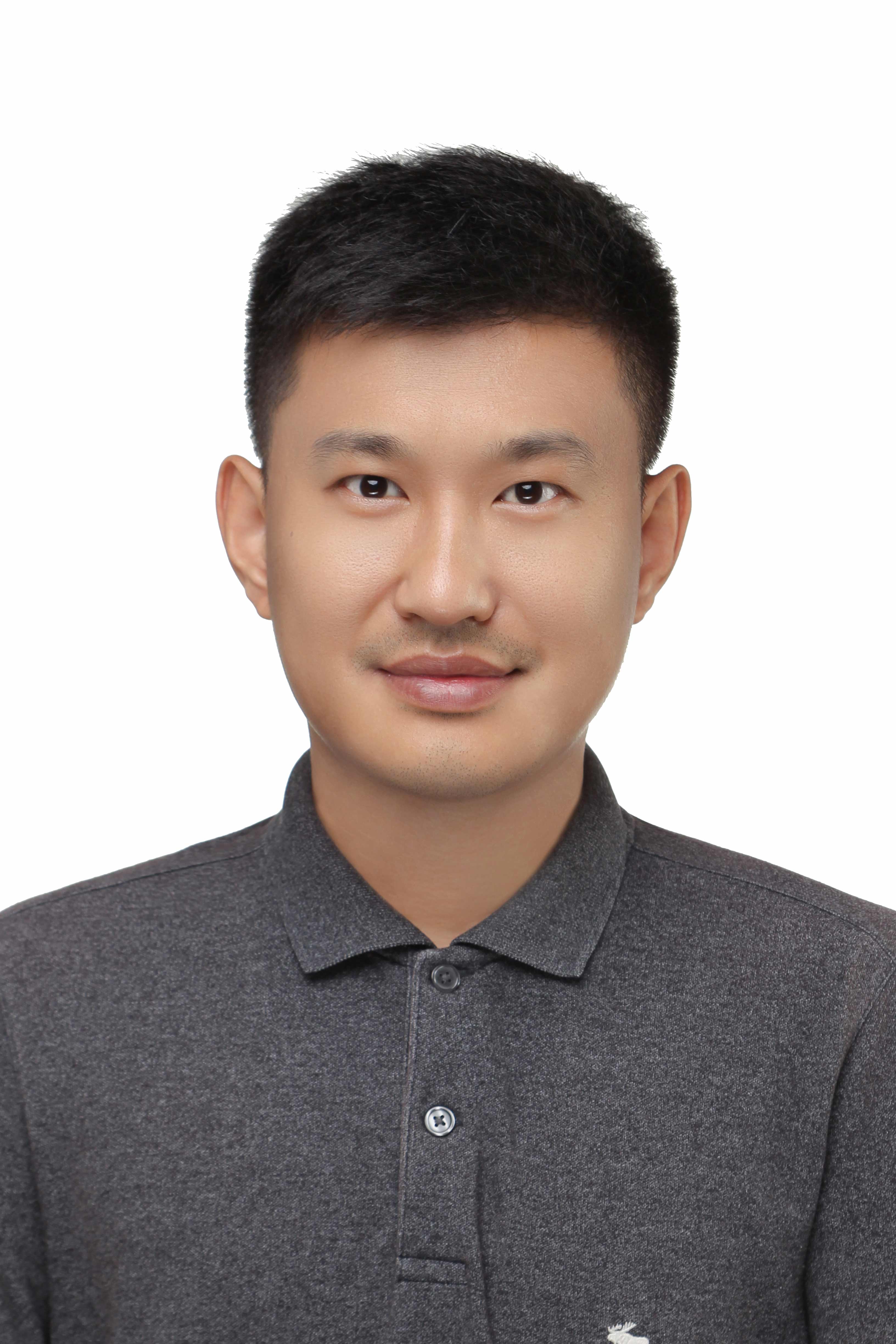}}]{Yong Du (Member, IEEE)} earned the B.Sc. and M.Sc. degrees from Jiangnan University, Wuxi, China, in 2011 and 2014, and the Ph.D. degree from South China University of Technology, Guangzhou, China, in 2019. He is currently an associate professor at the School of Computer Science and Technology, Ocean University of China. His research focuses on computer vision, image processing, machine learning, and generative models. He has published several papers in top venues, including CVPR, ICCV, ECCV, and IEEE Transactions on Image Processing, Multimedia, and Cybernetics.
\end{IEEEbiography}
\begin{IEEEbiography}[{\includegraphics[width=1in,height=1.25in,clip,keepaspectratio]{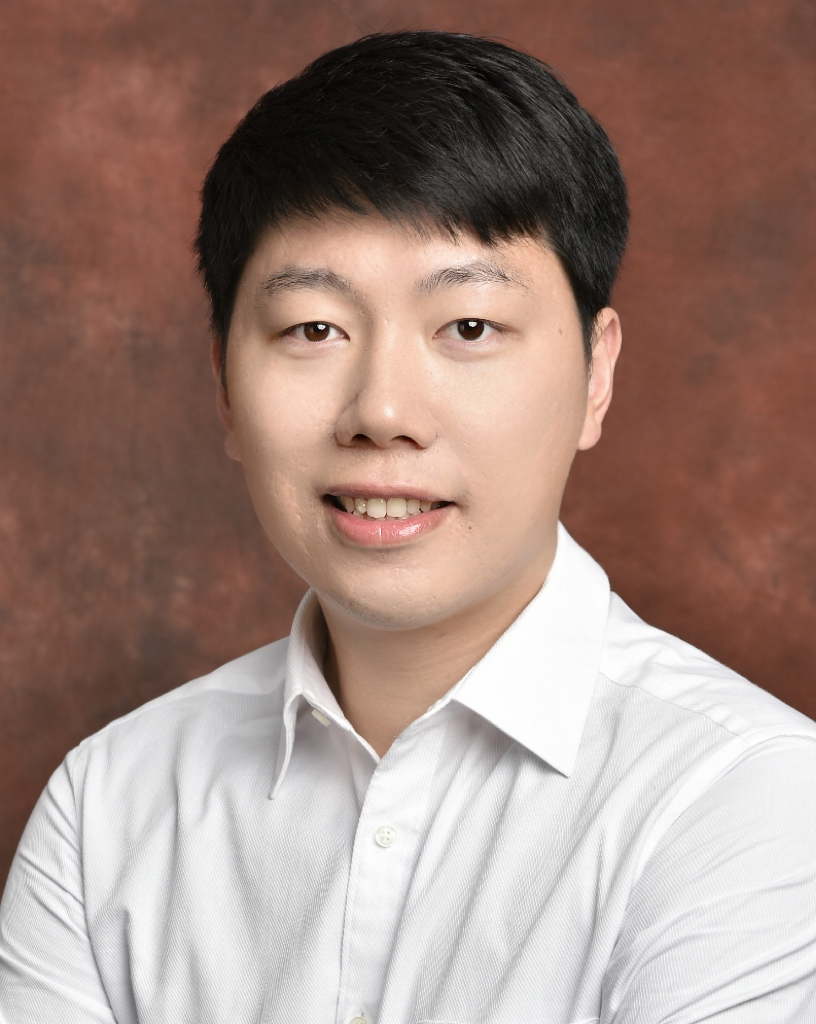}}]{Shengfeng He (Senior Member, IEEE)} is an associate professor in the School of Computing and Information Systems at Singapore Management University. He was a faculty member at South China University of Technology (2016–2022). He earned his B.Sc. and M.Sc. from Macau University of Science and Technology (2009, 2011) and a Ph.D. from City University of Hong Kong (2015). His research focuses on computer vision and generative models. He has received awards including the Google Research Award, PerCom 2024 Best Paper Award, and the Lee Kong Chian Fellowship. He is a senior IEEE member and a distinguished CCF member. He serves as lead guest editor for IJCV and associate editor for IEEE TNNLS, IEEE TCSVT, Visual Intelligence, and Neurocomputing. He is an area chair/senior PC member for ICML, AAAI, IJCAI, BMVC, and the Conference Chair of Pacific Graphics 2026.\end{IEEEbiography}

\end{document}